\documentclass[10pt,twocolumn,a4paper]{article}
\pdfoutput=1

\usepackage[pagenumbers]{cvpr} %

\usepackage[hmargin=2.25cm,vmargin=2.78cm]{geometry}

\usepackage{booktabs} %
\usepackage{multirow}

\usepackage{array}
\newcolumntype{L}[1]{>{\raggedright\let\newline\\\arraybackslash\hspace{0pt}}m{#1}}
\newcolumntype{C}[1]{>{\centering\let\newline\\\arraybackslash\hspace{0pt}}m{#1}}
\newcolumntype{R}[1]{>{\raggedleft\let\newline\\\arraybackslash\hspace{0pt}}m{#1}}

\usepackage{tikz}

\usepackage{adjustbox}

\usepackage{esvect}

\usepackage{algorithm}
\usepackage[noend]{algpseudocode} %
\floatname{algorithm}{Procedure}

\newcommand{\arghspace}{\:\:}

\usepackage{caption}
\usepackage{subcaption}

\usepackage{lipsum}

\newcommand{\sfm}{Structure-from-Motion}

\newcommand{\updateviewfeat}{\textsc{UpdateViewFeat}}
\newcommand{\updatescenepointfeat}{\textsc{UpdateScenePointFeat}}
\newcommand{\updateglobalfeat}{\textsc{UpdateGlobalFeat}}
\newcommand{\updateprojfeat}{\textsc{UpdateProjFeat}}
\newcommand{\graphcrossattn}{\textsc{GraphCrossAttention}}
\newcommand{\graphattnsfm}{\textsc{GraphAttentionSfm}}

\newcommand{\MORAN}{DPESFM}
\newcommand{\GESFM}{GESFM}
\newcommand{\LINEAR}{Linear}
\newcommand{\COLMAP}{Colmap}
\newcommand{\GPSFM}{GPSFM}
\newcommand{\PPSFM}{PPSFM}
\newcommand{\VARPRO}{VarPro}

\newcommand{\headerMORAN}{\MORAN{}}
\newcommand{\headerGESFM}{\GESFM{}}
\newcommand{\headerLINEAR}{\LINEAR{}}
\newcommand{\headerCOLMAP}{\COLMAP{}}
\newcommand{\headerGPSFM}{\GPSFM{}}
\newcommand{\headerPPSFM}{\PPSFM{}}
\newcommand{\headerVARPRO}{\VARPRO{}}

\newcommand{\sceneid}{}
\newcommand{\scenename}{}
\newcommand{\trimleft}{0.0}
\newcommand{\trimbottom}{0.0}
\newcommand{\trimright}{0.0}
\newcommand{\trimtop}{0.0}

\definecolor{cvprblue}{rgb}{0.21,0.49,0.74}
\usepackage[pagebackref,breaklinks,colorlinks,citecolor=cvprblue]{hyperref}
\usepackage{pgf,pgfarrows,pgfnodes}
\usepackage[accsupp]{axessibility} %

\def\paperID{8747} %
\def\confName{CVPR}
\def\confYear{2024}

\title{Learning Structure-from-Motion with Graph Attention Networks}

\author{{Lucas Brynte \qquad Jos\'e Pedro Iglesias \qquad Carl Olsson \qquad Fredrik Kahl}\\
Chalmers University of Technology\\
{\tt\small \{brynte,jose.iglesias,caols,fredrik.kahl\}@chalmers.se}
}

\begin{document}
\maketitle
\begin{abstract}
In this paper we tackle the problem of learning Structure-from-Motion (SfM) through the use of graph attention networks. SfM is a classic computer vision problem that is solved though iterative minimization of reprojection errors, referred to as Bundle Adjustment (BA), starting from a good initialization. In order to obtain a good enough initialization to BA, conventional methods rely on a sequence of sub-problems (such as pairwise pose estimation, pose averaging or triangulation) which provide an initial solution that can then be refined using BA. In this work we replace these sub-problems by learning a model that takes as input the 2D keypoints detected across multiple views, and outputs the corresponding camera poses and 3D keypoint coordinates. Our model takes advantage of graph neural networks to learn SfM-specific primitives, and we show that it can be used for fast inference of the reconstruction for new and unseen sequences. The experimental results show that the proposed model outperforms competing learning-based methods, and challenges COLMAP while having lower runtime. Our code is available at: \url{https://github.com/lucasbrynte/gasfm/}.
\end{abstract}

\section{Introduction}
\label{sec:intro}

 Structure-from-Motion (SfM) is a classic and still relevant problem in computer vision. The goal of SfM is to estimate camera poses and 3D coordinates of keypoints detected across multiple images, and can be formulated as an optimization over $m$ camera matrices $\{\mathbf{P}_i\}$, $i = 1,\hdots,m$ and $n$ 3D points $\{\mathbf{X}_j\}$ , $j=1,\hdots,n$ of the form
 \begin{equation}
\begin{aligned}
& \underset{\{\mathbf{P}_i\},\{\mathbf{X}_j\}}{\text{minimize}}
& &  \sum_{ij} r\left(\mathbf{m}_{ij},  \mathbf{z}_{ij} \right) \\
& \text{subject to}
& & \mathbf{z}_{ij} = \mathbf{P}_i \mathbf{\bar{X}}_j, \text{  } \forall i,j
\end{aligned}
\label{eq:sfm_optimization}
\end{equation}
where $\mathbf{m}_{ij}$ holds the 2D coordinates of the $j$\textsuperscript{th} keypoint in the $i$\textsuperscript{th} image. The loss in \eqref{eq:sfm_optimization} is generally chosen as the reprojection error 
\begin{equation}
    r\left(\mathbf{m}_{ij},  \mathbf{z}_{ij} \right) = || \mathbf{m}_{ij} - \Pi\left(\mathbf{z}_{ij}\right) ||^2
\end{equation}
where $\Pi(\mathbf{x}) = \left(\frac{x_1}{x_3},\frac{x_2}{x_3}\right)$, and the nonlinear least squares problem \eqref{eq:sfm_optimization}, referred to as Bundle Adjustment (BA)~\cite{hartley_zisserman_2004}, can be solved iteratively using second-order methods like Levenberg-Marquardt~\cite{hartley_zisserman_2004,triggs1999}. Given the sparsity of the problem, sparse computation methods~\cite{konolige2010} can be used in order to increase the efficiency of the optimization, allowing BA to be used even for scenes with a large number of views or points. However, it is widely known that BA is highly non-convex and tends to converge to the nearest local minimum when not initialized close to the globally optimal solution.
As a consequence, BA is typically the last step of a reconstruction pipeline, preceeding global SfM methods such as~\cite{ke-kanade-pami-2007,kahl-hartley-pami-2008,enqvist2011,eriksson-etal-pami-2021}, or incremental
SfM methods such as~\cite{agarwal2009,schoenberger2016sfm,schoenberger2016mvs}
that solve a sequence of subproblems like pairwise pose estimation, pose averaging, triangulation or camera resection~\cite{hartley1997,hartley_zisserman_2004,nister2004,kahl-etal-ijcv-2008}.
A different approach consists of projective factorization methods~\cite{dai2013,magerand2017,strum1996,zach2018,hong2017,iglesias2023} which factorize the $2m \times n$ measurement matrix into two rank four matrices corresponding to the camera matrices and 3D points (in homogeneous coordinates).
In particular, works like~\cite{zach2018,hong2017,iglesias2023} allow initialization-free SfM given their wide basin of convergence, meaning that their methods can be initialized with random camera poses and still converge with a high rate of success to the desired global minimum.
Even though these methods have been improving in terms of accuracy and robustness to missing data, factorization-based methods require the input data to be almost completely free of outliers which unfortunately cannot be guaranteed in most real world sequences or datasets, and hence severely compromises the usability of these methods.  

A common challenge with all these approaches to solve SfM is their scalability as the number of views and keypoints increase. Incremental SfM tries to tackle this issue by starting with a subset of the views, estimate its reconstruction and incrementally adding more views. Some factorization-based methods can also take advantage of the same sparse computation methods used in BA, which significantly improves their ability to scale with sequence size.
While this allows to reconstruct scenes with thousands of views and millions of points, it can still take hours to recover the reconstruction of a single scene. 

More recently, deep learning methods for 3D reconstruction and SfM have been proposed~\cite{mildenhall2020,wei2020deepsfm, chen2022, wei2021, Moran_2021_ICCV, wang2021,purkait2020, wang2023}. Given their increased complexity, these models are able to learn complex relations between the input data, outperforming conventional methods and achieving state-of-the-art results in 3D reconstruction and novel view synthesis when the camera pose and calibration are known~\cite{mildenhall2020,wei2020deepsfm, chen2022}. 
Works such as~\cite{wei2021, Moran_2021_ICCV, wei2020deepsfm} tackle SfM with deep learning models which different degrees of success. 

In DeepSfM~\cite{wei2020deepsfm} the authors propose an end-to-end approach to estimate camera pose and a dense depth map of a target view from multiple source views. Their method estimates a feature map for each of the source views, which are then combined with the feature map of the target view in order to estimate pose and depth cost volumes. These cost volumes are then fed to two heads that estimate the camera pose and depth map of the target view. 
In PoseDiffusion~\cite{wang2023}, a probabilistic diffusion framework is proposed to mirror the iterations of bundle adjustment, allowing it to incorporate geometric priors into the problem. The model is trained on object-centric scenes and it is shown that it can generalize to unseen scenes without further training. 

In~\cite{Moran_2021_ICCV}, which is the closest work to ours, the authors have pursued an approach similar to projective factorization, taking as input the 2D keypoints tracked along multiple views. %
These 2D point tracks are
processed by a number of
linear permutation-equivariant layers, i.e.
layers such that
a permutation
on the rows and/or columns of the input tensor results in the same permutation on the output.
The output of these layers is then fed to camera pose and 3D points coordinate
regression
heads.
The model is learned using reprojection errors as loss combined with a hinge loss term to push points
in front
of the cameras.
By using Adam~\cite{diederik2015} as optimizer along with gradient normalization, the authors show that their method
has improved convergence properties compared to random initializations.

A major benefit of learning-based methods is their
potential in
generalizing to new data after being trained. This can be particularly beneficial to SfM problems since occlusions and missing
measurements are common issues that make some scenes particularly challenging to reconstruct. In these cases, the learned model can act as a prior and help constrain the solution based on the data seen during training. However, most of the learning-based methods mentioned for 3D reconstruction or SfM require scene-specific optimization or training, which turns out to be particularly slow for large sequences. PoseDiffusion~\cite{wang2023} achieves some degree of generalization to unseen scenes, but it is conditional on the categories and object-centric datasets seen during training.
In~\cite{Moran_2021_ICCV} the authors actually show that their model can be trained on a large collection of scenes and then used for inference on unobserved scenes. However, without fine-tuning on these new scenes (which is time-consuming) the model's accuracy is significantly reduced.

In this paper we propose a graph attention network for initialization-free Structure-from-Motion that can be trained on multiple scenes and used for fast inference on new, unseen scenes. Similarly to~\cite{Moran_2021_ICCV}, our network takes as input sparse measured image points, matched across multiple viewpoints, which are processed through a sequence of permutation-equivariant graph cross attention layers.
The output of the graph cross attention layers is then processed by two regression heads, one for the 3D points and another for the camera poses.  By using graph attention layers instead of linear equivariant maps as in~\cite{Moran_2021_ICCV}, our model is able to better learn the implicit geometric primitives of SfM in the input data, leading to more expressiveness and better generalization to unseen sequences. Additionally, by not requiring time-consuming fine-tuning on unseen scenes, our model is able to provide fast inference, which can then be refined directly with BA with good convergence.

The contributions of the paper can be summarized as:
\begin{itemize}
    \item We propose graph attention networks for initialization-free Structure-from-Motion that estimates camera parameters and 3D coordinates of 2D keypoints tracked across multiple views;
    \item The method takes advantage of the graph cross-attention layers, which are equivariant to permutation on the input data regarding order of views and points. We show that these layers can model more complex relations between input data and when combined with data augmentation result in significantly improved performance when compared to baseline methods; 
    \item We evaluate the proposed learned method on unobserved scenes and show that it outperforms competing state-of-the art learning methods at inference in terms of accuracy without the need of scene-specific model fine-tuning; 
    \item We show that our method achieves competitive results when compared to state-of-the-art conventional Structure-from-Motion pipelines while having a significantly lower runtime. 
\end{itemize}

\nocite{kahl-cvpr-2001,hartley-kahl-ijcv-2007}

\section{Graph Attention Network Preliminaries}
\label{sec:graph_attn_nets}
Before introducing our proposed method, we will here briefly outline the graph attention networks that will be utilized -- the GATv2 model~\cite{brody2022how} based on~\cite{gatv1}.

The GATv2 model is a type of graph neural network (GNN) that carries out aggregation from neighboring nodes with a dynamic attention-based weighted average.
The input to a GATv2 layer is a set of current node features $\{\mathbf{h}_i\in \mathbb{R}^{d} \mid i\in \mathcal{V}\}$ along with a set of directed edges $\mathcal{E}$ defining the graph connectivity.
The layer outputs a new set of node features $\{\mathbf{h}'_i\in \mathbb{R}^{d'} \mid i\in \mathcal{V}\}$, by applying a single shared learned function on every node $h_i$ together with its neighborhood $\mathcal{N}_i=\{j \in \mathcal{V} \mid \left( j,i \right) \in \mathcal{E}\}$.

First, a scoring function $e: \mathbb{R}^{d}$$\,\times\,$$\mathbb{R}^{d}$$\,\rightarrow\,$$ \mathbb{R}$ is applied on all edges, by feeding the respective source and target node features as input to a shared 2-layer Multi-Layer Perceptron (MLP):
\begin{equation}
	e\left(\mathbf{h}_i, \mathbf{h}_j\right)=
	\mathbf{a}^{\top} \cdot \mathrm{LeakyReLU}
	\left(
		\mathbf{W} \left[\mathbf{h}_i \| \mathbf{h}_j\right]
	\right),
	\label{eq:gat}
\end{equation}
where $\mathbf{a}\in \mathbb{R}^{d'}$ and $\mathbf{W}\in \mathbb{R}^{d'\times 2d}$ are learned parameters, and $\|$ denotes vector concatenation.
The attention scores are then normalized across all neighbors:
\begin{equation}
	\alpha_{ij} =
	\mathrm{softmax}_j\left(
	e\left(\mathbf{h}_i, \mathbf{h}_{j}\right)
	\right) =
	\frac{\mathrm{exp}\left(e\left(\mathbf{h}_i, \mathbf{h}_j\right)\right)}{\sum\nolimits_{j'\in\mathcal{N}_i} \mathrm{exp}\left(e\left(\mathbf{h}_i, \mathbf{h}_{j'}\right)\right)},
	\label{eq:softmax}
\end{equation}
and the layer is completed by carrying out a weighted average of linear projections of all source node features:
\begin{equation}
	\mathbf{h}'_i=
		\sum\nolimits_{j\in\mathcal{N}_{i}}
		\alpha_{ij}
		\cdot\mathbf{W}\mathbf{h}_j
	.
	\label{eq:weighted_avg}
\end{equation}

\section{Method}
We apply a graph attention network model to perform \sfm{} on a desired scene based on its measured image point correspondences.
The model then outputs a reconstruction, represented by all $m$ camera matrices and all $n$ scene points.
Structure-from-Motion is in general a large and complex optimization problem, and it is not trivially addressed as a learning problem.
Our architecture combines the power and expressivity of the attention mechanism, and exploits its inherent permutation equivariance in a meaningful way, while balancing expressivity with efficiency, resulting in a very limited computational and memory footprint considering the task at hand.

The input data for a particular scene consists of a sparse set of measured image point correspondences (point tracks).
To represent this data, we define a binary observability matrix $\mathbf{O} \in \{0, 1\}^{m \times n}$, where $m$ is the number of views and $n$ is the number of scene points, and where $\mathbf{O}_{ij} = 1$ if and only if scene point $j$ is observed in image $i$.
Furthermore, let
$\mathcal{P} = \{ \mathbf{m}_{ij} \in \mathbb{R}^2 \mid \mathbf{O}_{ij} = 1 \}$
be the collection of observed image point tracks, where $\mathbf{m}_{ij}$ is the (normalized) measured image point for the projection of scene point $j$ in camera $i$.

\subsection{Graph Attention Network Architecture}
\label{sec:arch}
In this section, we will outline our graph attention network architecture, starting by introducing our feature representation.
Then we proceed to define update operations for different feature types, and finally define the entire network architecture based on these operations.

\begin{figure}
    \centering
    \input{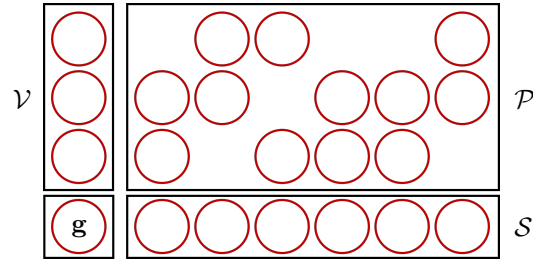}
    \caption{Illustration of the different types of features in the network architecture, for $m=3$ cameras and $n=6$ scene points.
    The projection features $\mathcal{P}$ are represented by a sparse vector-valued matrix, with its sparsity pattern determined by the point track measurements, while view features $\mathcal{V}$ and scene point features $\mathcal{S}$ are represented by dense vector-values column and row vectors.
    A single vector $\mathbf{g}$ holds global features.
    }
    \label{fig:features}
\end{figure}

\paragraph{Feature representation. }
Throughout the network architecture, we maintain and continuously update a set of features of various types\footnote{For simplicity, we use a slight abuse of set notation here. If there are feature vector duplicates, they are still considered as separate elements.}:

\begin{itemize}
    \item \emph{Projection features},
        $\mathcal{P} = \{ \mathbf{p}_{ij} \in \mathbb{R}^{d_p} \mid \mathbf{O}_{ij} = 1 \}$.
    \item \emph{View features},
        $\mathcal{V} = \{ \mathbf{v}_i \in \mathbb{R}^{d_v} \mid i=1,2,\dots,m \}$.
    \item \emph{Scene point features},
    $\mathcal{S} = \{ \mathbf{s}_j \in \mathbb{R}^{d_s} \mid j=1,2,\dots,n \}$.
    \item \emph{Global features},
        $\mathbf{g} \in \mathbb{R}^{d_g}$.
\end{itemize}
The \emph{projection features} $\mathbf{p}_{ij}$ are feature vectors specific to every observed image point.
In practice, these are bundled together and organized in a sparse vector-valued matrix, as illustrated by the large top-right matrix in Figure~\ref{fig:features}.
The sparsity pattern of $\mathcal{P}$ will be fixed throughout the network and is determined by $\mathbf{O}$, which however depends on the scene.
The \emph{view features} $\mathcal{V}$ and \emph{scene point features} $\mathcal{S}$, on their part, are organized in dense vector-valued column and row vectors, respectively, also illustrated in Figure~\ref{fig:features}.
Finally, the \emph{global features} $\mathbf{g}$, consist of a single globally shared feature vector, and is illustrated in the bottom-left corner of Figure~\ref{fig:features}.

In order to define graph operations,
we see every feature vector as being associated with a corresponding graph node.
At this point, note that a simple approach would be to connect all nodes
$\mathcal{P}$, $\mathcal{V}$, $\mathcal{S}$, and $\{ \mathbf{g} \}$
in one unified graph, and directly apply a graph neural network with global weight sharing.
Instead, since the nodes are semantically different, we consider these semantics when choosing the level of weight sharing, and even vary the feature dimension depending on the type of node.
In practice, we
construct different graphs for
propagation between different types of nodes.

\newcommand{\relu}{\textrm{ReLU}}
\newcommand{\layernorm}{\textrm{LN}}
\newcommand{\linear}{\textrm{Linear}}
\newcommand{\ffn}{\textrm{FFN}}
\newcommand{\emptyarg}{-}

Having defined the different feature types, we proceed to define corresponding update operations, which are repeatedly carried out throughout the network.
In all pseudo-code,
\relu{} denotes to the rectified linear unit and \layernorm{} denotes layer normalization~\cite{ba2016layer}.
\linear{} denotes a learned affine layer and \ffn{} denotes a number of stacked \linear{} layers interleaved with \relu{} activations.
Subscripts may occur to emphasize feature dimensions.
Note that all of these operations may be applied on multiple nodes at once, with shared parameters.
The $+$ notation implies node-wise addition.

\paragraph{Graph Cross-Attention. }
Procedure~\ref{alg:graphcrossattn} is utilized for all feature aggregations throughout the network.
It carries out cross-attention~\cite{vaswani_nips2017}, attending to and propagating information from one set of input / source nodes $\mathcal{H}_1$ to another set of output / target nodes $\widehat{\mathcal{H}}_2$.
The previous target node features $\mathcal{H}_2$, if provided, will act as query features when computing attention scores.
In the first few layers these may not be available, and initial zero-features are used instead.

In our context, the source nodes and target nodes correspond to different feature types, possibly with different feature dimension.
As the source and target nodes are disjoint, we may represent the connectivity by a directed bipartite graph.
Furthermore, in our context this graph will often consist of a number of disconnected subgraphs.
Procedure\ref{alg:graphcrossattn} essentially acts as a wrapper around a GATv2 layer~\cite{brody2022how} (for which we use the PyTorch Geometric~\cite{pytorch-geometric} implementation), while carrying out input normalization, and feature projections as necessary.
We consistently use 4 attention heads for the GATv2 layers.
\begin{algorithm}
    \caption{\graphcrossattn{}}
    \label{alg:graphcrossattn}
    \footnotesize
    \begin{algorithmic}[1] %
        \Require $\mathcal{H}_1 \arghspace{}$ Source node features.
        \Require $\mathcal{H}_2 \arghspace{}$ Previous target node features (optional).
        \Require $\mathcal{E}_{\mathcal{H}_1 \to \mathcal{H}_2} \arghspace{}$ Set of edges from $\mathcal{H}_1 \to \mathcal{H}_2 \arghspace{}$.
        \Require $d_1 \arghspace{}$ Source node feature dimension.
        \Require $d_2 \arghspace{}$ Target node feature dimension.
        \Ensure $\widehat{\mathcal{H}}_2 \arghspace{}$ Updated target node features.

        \State $\widehat{\mathcal{H}}_1 \gets \relu( \layernorm( \mathcal{H}_1 ) )$
        \If{$\mathcal{H}_2$ provided}
            \State $\widehat{\mathcal{H}}_2 \gets \relu( \layernorm( \mathcal{H}_2 ) )$
            \If{$d_1 \neq d_2$}
                \State $\widehat{\mathcal{H}}_2 \gets \linear{}_{d_2 \to d_1} ( \widehat{\mathcal{H}}_2 )$
            \EndIf
        \Else
            \State $\widehat{\mathcal{H}}_2 \gets $ zero features $\vv{0} \in \mathbb{R}^{d_1}$
        \EndIf
        \State $\widehat{\mathcal{H}}_2 \gets \textrm{GATv2Conv} ( \widehat{\mathcal{H}}_1, \widehat{\mathcal{H}}_2, \mathcal{E}_{\mathcal{H}_1 \to \mathcal{H}_2} )$
        \If{$d_1 \neq d_2$}
            \State $\widehat{\mathcal{H}}_2 \gets \linear{}_{d_1 \to d_2} ( \widehat{\mathcal{H}}_2 )$
        \EndIf
    \end{algorithmic}
\end{algorithm}

\begin{figure}
    \centering
    \input{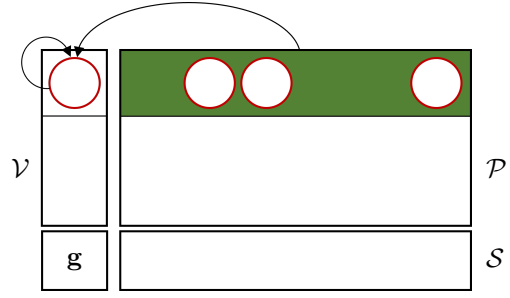}
    \caption{Illustration of the update of a single view feature.
    All $\mathcal{V}$ features are updated based on their previous
    value
    and the corresponding rows of $\mathcal{P}$.
    }
    \label{fig:updateviewfeat}
\end{figure}

\begin{figure}
    \centering
    \input{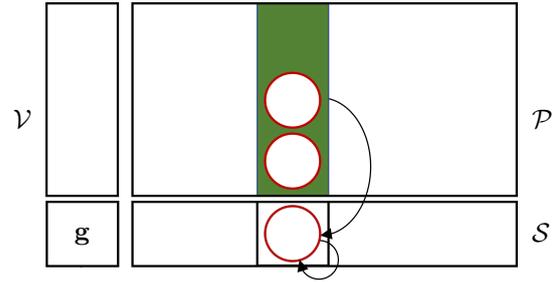}
    \caption{Illustration of the update of a single scene point feature.
    All $\mathcal{S}$ features are updated based on their previous
    value
    and the corresponding columns of $\mathcal{P}$.
    }
    \label{fig:updatescenepointfeat}
\end{figure}

\paragraph{View \& Scene Point Feature Update. }
View features and scene point features are updated by Procedure~\ref{alg:updateviewfeat} and~\ref{alg:updatescenepointfeat}, respectively, entirely symmetrical to one another.
A directed bipartite graph is constructed by taking $\mathcal{P}$ as source nodes, and $\mathcal{V}$ (or $\mathcal{S}$) as target nodes, and connecting every projection node with its corresponding view (or scene point) node, corresponding to rows (or columns) of $\mathcal{P}$.
Graph cross-attention according to Procedure~\ref{alg:graphcrossattn} is then carried out on the graph, and used as a residual mapping.
This is followed by another residual mapping with a feed-forward network.
Illustrations of these update operations are provided in Figures~\ref{fig:updateviewfeat} and~\ref{fig:updatescenepointfeat}.
\begin{algorithm}
    \caption{\updateviewfeat{}}
    \label{alg:updateviewfeat}
    \footnotesize
    \begin{algorithmic}[1] %
        \Require $\mathcal{P}, \mathbf{O} \arghspace{}$ Projection features + observability matrix.
        \Require $\mathcal{V} \arghspace{}$ Previous view features (optional).
        \Require $d_p, d_v \arghspace{}$ Feature dimensions.
        \Ensure $\widehat{\mathcal{V}} \arghspace{}$ Updated view features.

        \State $\mathcal{E}_{\mathcal{P} \to \mathcal{V}} \gets \{ (\mathcal{P}_{ij}, \mathcal{V}_i) \mid \mathbf{O}_{ij} = 1,\;\; i=1,2,\dots,m \}$
        \If{$\mathcal{V}$ provided}
            \State $\widehat{\mathcal{V}} \gets \mathcal{V} + \graphcrossattn{}(\mathcal{P}, \mathcal{V}, \mathcal{E}_{\mathcal{P} \to \mathcal{V}}, d_p, d_v)$
        \Else
            \State $\widehat{\mathcal{V}} \gets \graphcrossattn{}(\mathcal{P}, \mathcal{E}_{\mathcal{P} \to \mathcal{V}}, d_p, d_v)$
        \EndIf
        \State $\widehat{\mathcal{V}} \gets \widehat{\mathcal{V}} + \ffn{}( \relu{}( \layernorm( \widehat{\mathcal{V}} ) ) )$
    \end{algorithmic}
\end{algorithm}
\begin{algorithm}
    \caption{\updatescenepointfeat{}}
    \label{alg:updatescenepointfeat}
    \footnotesize
    \begin{algorithmic}[1] %
        \Require $\mathcal{P}, \mathbf{O} \arghspace{}$ Projection features + observability matrix.
        \Require $\mathcal{S} \arghspace{}$ Previous scene point features (optional).
        \Require $d_p, d_s \arghspace{}$ Feature dimensions.
        \Ensure $\widehat{\mathcal{S}} \arghspace{}$ Updated scene point features.

        \State $\mathcal{E}_{\mathcal{P} \to \mathcal{S}} \gets \{ (\mathcal{P}_{ij}, \mathcal{S}_i) \mid \mathbf{O}_{ij} = 1,\;\; i=1,2,\dots,m \}$
        \If{$\mathcal{S}$ provided}
            \State $\widehat{\mathcal{S}} \gets \mathcal{S} + \graphcrossattn{}(\mathcal{P}, \mathcal{S}, \mathcal{E}_{\mathcal{P} \to \mathcal{S}}, d_p, d_s)$
        \Else
            \State $\widehat{\mathcal{S}} \gets \graphcrossattn{}(\mathcal{P}, \mathcal{E}_{\mathcal{P} \to \mathcal{S}}, d_p, d_s)$
        \EndIf
        \State $\widehat{\mathcal{S}} \gets \widehat{\mathcal{S}} + \ffn{}( \relu{}( \layernorm( \widehat{\mathcal{S}} ) ) )$
    \end{algorithmic}
\end{algorithm}

\begin{figure}
    \centering
    \input{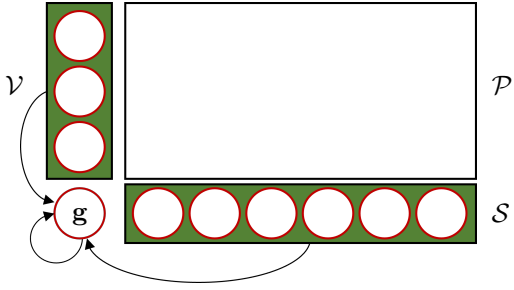}
    \caption{Illustration of the global feature update, where $\mathbf{g}$ is updated based on its previous value, together with an aggregation of all $\mathcal{V}$ and $\mathcal{S}$ features.
    }
    \label{fig:updateglobalfeat}
\end{figure}

\paragraph{Global Feature Update. }
The global feature update, outlined in Procedure~\ref{alg:updateglobalfeat} is carried out in almost the same manner, but now aggregating from all view features $\mathcal{V}$ and scene point features $\mathcal{S}$ instead of $\mathcal{P}$.
This is carried out with two independent aggregations, with different learned parameters, and the sum of both aggregations constitute the residual mapping.
Figure~\ref{fig:updateglobalfeat} provides an illustration for the update operation.
\begin{algorithm}
    \caption{\updateglobalfeat{}}
    \label{alg:updateglobalfeat}
    \footnotesize
    \begin{algorithmic}[1] %
        \Require $\mathcal{V} \arghspace{}$ View features.
        \Require $\mathcal{S} \arghspace{}$ Scene point features.
        \Require $\mathbf{g} \arghspace{}$ Previous global features (optional).
        \Require $d_v, d_s, d_g \arghspace{}$ Feature dimensions.
        \Ensure $\widehat{\mathbf{g}} \arghspace{}$ Updated global features.

        \State $\mathcal{E}_{\mathcal{V} \to \mathbf{g}} \gets \{ (\mathcal{V}_{i}, \mathbf{g}) \mid i=1,2,\dots,m \}$
        \State $\mathcal{E}_{\mathcal{S} \to \mathbf{g}} \gets \{ (\mathcal{S}_{j}, \mathbf{g}) \mid j=1,2,\dots,n \}$
        \If{$\mathbf{g}$ provided}
            \State $\widehat{\mathbf{g}}_v \gets \graphcrossattn{}(\mathcal{V}, \{\mathbf{g}\}, \mathcal{E}_{\mathcal{V} \to \mathbf{g}}, d_v, d_g)$
            \State $\widehat{\mathbf{g}}_s \gets \graphcrossattn{}(\mathcal{S}, \{\mathbf{g}\}, \mathcal{E}_{\mathcal{S} \to \mathbf{g}}, d_s, d_g)$
            \State $\widehat{\mathbf{g}} \gets \mathbf{g} + \widehat{\mathbf{g}}_v + \widehat{\mathbf{g}}_s$
        \Else
            \State $\widehat{\mathbf{g}}_v \gets \graphcrossattn{}(\mathcal{V}, \emptyarg{}, \mathcal{E}_{\mathcal{V} \to \mathbf{g}}, d_v, d_g)$
            \State $\widehat{\mathbf{g}}_s \gets + \graphcrossattn{}(\mathcal{S}, \emptyarg{}, \mathcal{E}_{\mathcal{S} \to \mathbf{g}}, d_s, d_g)$
            \State $\widehat{\mathbf{g}} \gets \widehat{\mathbf{g}}_v + \widehat{\mathbf{g}}_s$
        \EndIf
        \State $\widehat{\mathbf{g}} \gets \widehat{\mathbf{g}} + \ffn{}( \relu{}( \layernorm( \widehat{\mathbf{g}} ) ) )$
    \end{algorithmic}
\end{algorithm}

\begin{figure}
    \centering

    \input{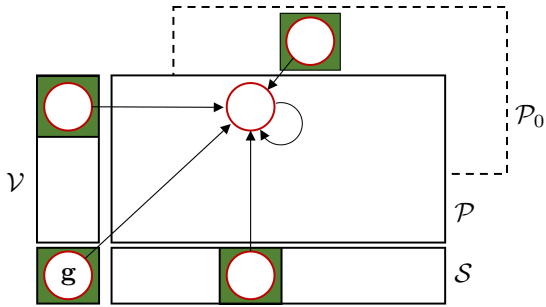}
    
    \caption{Illustration of the update of a single projection feature.
    All $\mathcal{P}$ features are updated based on their previous value, as well as initial projection features $\mathcal{P}_0$, the global feature $\mathbf{g}$, and the corresponding $\mathcal{V}$ and $\mathcal{S}$ features.
    }
    \label{fig:updateprojfeat}
\end{figure}

\paragraph{Projection Feature Update. }
The projection features are updated according to Procedure~\ref{alg:updateprojfeat}.
No aggregation is involved during this step.
Instead, at every projection node the corresponding $\mathcal{V}$, $\mathcal{S}$ and $\mathbf{g}$ features are all collected along with the current projection features $\mathcal{P}$, as illustrated in Figure~\ref{fig:updateprojfeat}.
Each of these sources then undergoes a \relu{} activation and normalization with \layernorm{}, before being concatenated.
Note that the initial projection features $\mathcal{P}_0$ are typically also added to the input, and in this case they are simply concatenated to the previous projection features before the procedure is called.
Finally, a residual mapping is applied, by feeding the concatenated features to a shared feed-forward network.
\begin{algorithm}
    \caption{\updateprojfeat{}}
    \label{alg:updateprojfeat}
    \footnotesize
    \begin{algorithmic}[1] %
        \Require $\mathcal{P} \arghspace{}$ Previous projection features.
        \Require $\mathcal{V} \arghspace{}$ View features.
        \Require $\mathcal{S} \arghspace{}$ Scene point features.
        \Require $\mathbf{g} \arghspace{}$ Global features.
        \Require $d_p^{in}, d_v, d_s, d_g, d_p^{out} \arghspace{}$ Feature dimensions.
        \Ensure $\widehat{\mathcal{P}} \arghspace{}$ Updated projection features.

        \State $\widehat{\mathcal{P}} \gets \relu{}(\layernorm{}(\mathcal{V})) \;\|\; \relu{}(\layernorm{}(\mathcal{S}))$
            \Statex \hspace{\algorithmicindent} $\|\; \relu{}(\layernorm{}(\mathbf{g})) \;\|\; \relu{}(\layernorm{}(\mathcal{P}))$
        \Comment{View, scene point, and global features are implicitly distributed to their corresponding projection features before feature concatenation.}
        \State $\widehat{\mathcal{P}} \gets \mathcal{P} + \ffn{}_{(d_v+d_s+d_g+d_p^{in}) \to d_p^{out}}( \widehat{\mathcal{P}} )$

    \end{algorithmic}
\end{algorithm}

\paragraph{Entire Network Architecture. }
Now we have all the pieces to define the complete network architecture, which is outlined in Procedure~\ref{alg:graphattnsfm}.
The structure is straightforward:
1) \updateviewfeat{} and \updatescenepointfeat{} (Procedures~\ref{alg:updateviewfeat} and~\ref{alg:updatescenepointfeat}) are called to aggregate projection features.
2) \updateglobalfeat{} (Procedure~\ref{alg:updateglobalfeat}) is called to aggregate the resulting features further, to a single global feature vector.
3) \updateprojfeat{} (Procedure~\ref{alg:updateprojfeat}) is called in order to again distribute the aggregated information across the projection features.
These steps are repeated $L$ times, where we refer to $L$ as the number of layers.
Then, a final call to \updateviewfeat{} and \updatescenepointfeat{} is carried out, and the resulting view and scene point features are further processed by two corresponding regression heads, consisting of shared 3-layer feed-forward networks (\ffn{}) with hidden neurons of the same dimension as the input ($d_v$ or $d_s$).
The scene point head simply outputs the 3 coordinates of every estimated scene point.
For Euclidean reconstruction, the view head outputs the 3 coordinates of the location of the camera center, together with a unit-norm quaternion representing its orientation.
For projective reconstruction, all 12 camera matrix elements are regressed, followed by a normalization in the same manner as~\cite{Moran_2021_ICCV}.
Except for the regression heads, we use 1-layer \ffn{}s throughout the network, which in general
seemed slightly easier to train.

\begin{algorithm}
    \caption{\graphattnsfm{}}
    \label{alg:graphattnsfm}
    \footnotesize
    \begin{algorithmic}[1] %
        \Require $\mathcal{M} \arghspace{}$ Image point tracks, normalized.
        \Require $\mathbf{O} \arghspace{}$ Observability matrix.
        \Require $d_p, d_v, d_s, d_g \arghspace{}$ Feature dimensions.
        \Require $L \arghspace{}$ Number of layers.
        \Ensure $\mathcal{V} \arghspace{}$ Estimated camera parameters.
        \Ensure $\mathcal{S} \arghspace{}$ Estimated scene points.

        \State $\mathcal{P}_0 \gets \linear{}_{2 \to 2}( \mathcal{M} )$
        \Comment{Initial projection features: linear embedding of image points.}

        \State $\mathcal{V} \gets \updateviewfeat{}(\mathcal{P}_0, \mathbf{O}, \emptyarg{}, 2, d_v)$
        \State $\mathcal{S} \gets \updatescenepointfeat{}(\mathcal{P}_0, \mathbf{O}, \emptyarg{}, 2, d_s)$
        \State $\mathbf{g} \gets \updateglobalfeat{}(\mathcal{V}, \mathcal{S}, \emptyarg{}, d_v, d_s, d_g)$

        \For{$l = 1,2,\dots,L$}
            \If{$l=1$}
                \State $\mathcal{P} \gets \updateprojfeat{}( \mathcal{P}_0, \mathcal{V}, \mathcal{S}, \mathbf{g}, 2, d_v, d_s, d_g, d_p )$
            \Else
                \State $\mathcal{P} \gets \updateprojfeat{}( \mathcal{P} \| \mathcal{P}_0, \mathcal{V}, \mathcal{S}, \mathbf{g},$\\\hspace{8.2ex}
                $d_p+2, d_v, d_s, d_g, d_p )$
            \EndIf
            \State $\mathcal{V} \gets \updateviewfeat{}(\mathcal{P}, \mathbf{O}, \mathcal{V}, d_p, d_v)$
            \State $\mathcal{S} \gets \updatescenepointfeat{}(\mathcal{P}, \mathbf{O}, \mathcal{S}, d_p, d_s)$
            \If{$l < L$}
                \State $\mathbf{g} \gets \updateglobalfeat{}(\mathcal{V}, \mathcal{S}, \mathbf{g}, d_v, d_s, d_g)$
                \EndIf
        \EndFor
        \State $\{\mathbf{P}_i\} \gets \ffn{}( \relu{}( \mathcal{V} ) )$
        \Comment{Camera regression head}
        \State $\{\mathbf{X}_j\} \gets \ffn{}( \relu{}( \mathcal{S} ) )$
        \Comment{Scene point regression head}
    \end{algorithmic}
\end{algorithm}

\subsection{Loss Function}
The network is trained using the Adam optimizer, and as loss function we use the (non-squared) reprojection error, averaged over all observed image points.
To handle negative depths and overcome the singularity at the principal plane, we substitute the reprojection error with minus the depth in the loss function for any projection with depth smaller than a threshold $h=1e-4$.
We also apply gradient normalization in the same manner as~\cite{Moran_2021_ICCV}.

\subsection{Data Augmentation}\label{sec:data_aug}
For Euclidean reconstruction, where the image points are normalized using the intrinsic camera parameters, and therefore geometrically interpretable, we experiment with augmenting the training scenes with slightly transformed variations.
While the scene points remain fixed, each camera is randomly rotated about its center, first by a uniformly sampled angle $\alpha \in [-15, 15]$ about the principal axis, followed by a uniformly sampled angle $\gamma \in [-20, 20]$ about an axis orthogonal to the principal axis with a random direction.
The image points are transformed accordingly.

\subsection{Artificial Outlier Injection}\label{sec:outlier_injection}
We also consider the presence of measurement outliers, by artificially corrupting the data.
This is done by: 1) Selecting a subset of the 2D keypoint measurements to be replaced by artificial outliers. 2) For each view, replace all measurements marked to be outliers with random samples from a bivariate normal distribution, which is fit to all remaining (inlier) points.
For step 1), we carefully select points for outlier injection such that we avoid getting any scene points visible in $<2$ views or any views with $<8$ scene points visible, if regarding only the inlier projections.
In the supplementary material, the details are given on how this is achieved.
Note that while the network input is corrupted by outlier injection, the uncorrupted data is still used as learning targets for the loss function.

\section{Results}
\begin{table*}[t]
    \centering
    \footnotesize

\resizebox{\textwidth}{!}{%
\begin{tabular}{L{0.13cm}l | rr | rr | rr | rr | r}
 &
    & \multicolumn{4}{c|}{Without data augmentation}
    & \multicolumn{4}{c|}{With data augmentation}
    &  \\

 &
    & \multicolumn{2}{c|}{Inference} & \multicolumn{2}{c|}{Inference + BA}
    & \multicolumn{2}{c|}{Inference} & \multicolumn{2}{c|}{Inference + BA}
    &  \\

 &
    & Ours & \headerMORAN{}
    & Ours & \headerMORAN{}
    & Ours & \headerMORAN{}
    & Ours & \headerMORAN{}
    & \headerCOLMAP{} \\

\hline

\multirow{11}{0.13cm}{\rotatebox{90}{Reprojection error (px)}}
& Alcatraz Courtyard
    & 68.41 & 68.50 %
    & 1.70 & \textbf{0.81} (0.82) %
    & 36.01 & 92.37 %
    & \textbf{0.81} & 0.92 %
    & \textbf{0.81} \\ %
& Alcatraz Water Tower
    & 36.30 & 50.47 %
    & 1.05 & 1.13 (\textbf{0.55}) %
    & 87.67 & 2831.94 %
    & 0.88 & 10.16 %
    & \textbf{0.55} \\ %
& Drinking Fountain Somewhere in Zurich
    & 36.02 & 45.87 %
    & 0.55 & \textbf{0.31} (7.21) %
    & 219.75 & 234.90 %
    & \textbf{0.31} & 6.73 %
    & \textbf{0.31} \\ %
& Nijo Castle Gate
    & 65.71 & 64.53 %
    & \textbf{0.73} & \textbf{0.73} (5.81) %
    & 61.41 & 68.19 %
    & 0.88 & 0.89 %
    & \textbf{0.73} \\ %
& Porta San Donato Bologna
    & 74.32 & 94.26 %
    & \textbf{0.74} & \textbf{0.74} (1.10) %
    & 52.15 & 84.46 %
    & 0.76 & 0.75 %
    & 0.75 \\ %
& Round Church Cambridge
    & 61.18 & 55.51 %
    & \textbf{0.39} & \textbf{0.39} (0.50) %
    & 29.80 & 59.54 %
    & \textbf{0.39} & 1.49 %
    & \textbf{0.39} \\ %
& Smolny Cathedral St Petersburg
    & 156.16 & 120.85 %
    & \textbf{0.81} & \textbf{0.81} (15.15) %
    & 85.38 & 87.81 %
    & \textbf{0.81} & \textbf{0.81} %
    & \textbf{0.81} \\ %
& Some Cathedral in Barcelona
    & 150.01 & 146.10 %
    & 10.42 & 12.71 (21.46) %
    & 125.68 & 687.83 %
    & \textbf{0.89} & 16.77 %
    & \textbf{0.89} \\ %
& Sri Veeramakaliamman Singapore
    & 121.02 & 157.39 %
    & 2.19 & 16.87 (16.92) %
    & 83.50 & 166.68 %
    & 2.13 & 9.30 %
    & \textbf{0.71} \\ %
& Yueh Hai Ching Temple Singapore
    & 37.46 & 52.59 %
    & \textbf{0.65} & \textbf{0.65} (1.16) %
    & 25.60 & 51.35 %
    & \textbf{0.65} & 0.73 %
    & \textbf{0.65} \\ %
& Average
    & 80.66 & 85.61 %
    & 1.92 & 3.52 (7.07) %
    & 80.69 & 436.51 %
    & 0.85 & 4.86 %
    & \textbf{0.66} \\ %

\hline

\multirow{11}{0.13cm}{\rotatebox{90}{Rotation error (deg)}}
& Alcatraz Courtyard
    & 10.102 & 13.201 %
    & 2.607 & 0.035 %
    & 6.093 & 10.946 %
    & 0.038 & \textbf{0.030} %
    & 0.043 \\ %
& Alcatraz Water Tower
    & 10.637 & 11.053 %
    & 0.499 & 0.764 %
    & 11.501 & 10.641 %
    & 0.699 & 19.351 %
    & \textbf{0.228} \\ %
& Drinking Fountain Somewhere in Zurich
    & 15.846 & 16.014 %
    & 0.003 & \textbf{0.001} %
    & 15.415 & 15.704 %
    & \textbf{0.001} & 22.759 %
    & 0.007 \\ %
& Nijo Castle Gate
    & 16.751 & 10.546 %
    & 0.062 & 0.062 %
    & 17.347 & 20.032 %
    & 0.038 & \textbf{0.036} %
    & 0.064 \\ %
& Porta San Donato Bologna
    & 23.839 & 24.120 %
    & 0.095 & \textbf{0.094} %
    & 18.411 & 25.004 %
    & \textbf{0.094} & \textbf{0.094} %
    & 0.099 \\ %
& Round Church Cambridge
    & 18.906 & 14.473 %
    & 0.029 & \textbf{0.026} %
    & 10.295 & 18.685 %
    & 0.029 & 1.086 %
    & 0.035 \\ %
& Smolny Cathedral St Petersburg
    & 19.387 & 17.971 %
    & 0.023 & 0.022 %
    & 11.662 & 14.380 %
    & 0.023 & \textbf{0.019} %
    & 0.029 \\ %
& Some Cathedral in Barcelona
    & 27.270 & 30.471 %
    & 10.009 & 20.050 %
    & 27.908 & 29.119 %
    & \textbf{0.020} & 47.892 %
    & 0.025 \\ %
& Sri Veeramakaliamman Singapore
    & 28.275 & 36.903 %
    & 0.549 & 4.871 %
    & 23.702 & 36.176 %
    & 0.457 & 2.759 %
    & \textbf{0.169} \\ %
& Yueh Hai Ching Temple Singapore
    & 15.733 & 22.706 %
    & \textbf{0.038} & \textbf{0.038} %
    & 9.515 & 21.561 %
    & \textbf{0.038} & \textbf{0.038} %
    & 0.043 \\ %
& Average
    & 18.675 & 19.746 %
    & 1.391 & 2.596 %
    & 15.185 & 20.225 %
    & 0.144 & 9.406 %
    & \textbf{0.074} \\ %

\hline

\multirow{11}{0.13cm}{\rotatebox{90}{Translation error (m)}}
& Alcatraz Courtyard
    & 4.82 & 4.93 %
    & 1.09 & \textbf{0.01} %
    & 2.73 & 5.74 %
    & \textbf{0.01} & \textbf{0.01} %
    & \textbf{0.01} \\ %
& Alcatraz Water Tower
    & 8.66 & 7.35 %
    & 0.31 & 0.44 %
    & 7.53 & 7.77 %
    & 0.41 & 9.05 %
    & \textbf{0.12} \\ %
& Drinking Fountain Somewhere in Zurich
    & 4.44 & 4.44 %
    & \textbf{0.00} & \textbf{0.00} %
    & 4.45 & 4.46 %
    & \textbf{0.00} & 1.38 %
    & \textbf{0.00} \\ %
& Nijo Castle Gate
    & 5.07 & 3.08 %
    & \textbf{0.01} & \textbf{0.01} %
    & 5.67 & 6.95 %
    & \textbf{0.01} & \textbf{0.01} %
    & \textbf{0.01} \\ %
& Porta San Donato Bologna
    & 9.50 & 10.72 %
    & \textbf{0.05} & \textbf{0.05} %
    & 4.80 & 10.48 %
    & \textbf{0.05} & \textbf{0.05} %
    & \textbf{0.05} \\ %
& Round Church Cambridge
    & 8.94 & 7.19 %
    & \textbf{0.01} & \textbf{0.01} %
    & 5.28 & 9.00 %
    & \textbf{0.01} & 0.56 %
    & \textbf{0.01} \\ %
& Smolny Cathedral St Petersburg
    & 2.70 & 2.43 %
    & \textbf{0.01} & \textbf{0.01} %
    & 2.33 & 2.52 %
    & \textbf{0.01} & \textbf{0.01} %
    & \textbf{0.01} \\ %
& Some Cathedral in Barcelona
    & 12.64 & 12.69 %
    & 3.22 & 6.38 %
    & 12.32 & 12.66 %
    & \textbf{0.01} & 11.93 %
    & \textbf{0.01} \\ %
& Sri Veeramakaliamman Singapore
    & 4.94 & 4.90 %
    & 0.16 & 1.32 %
    & 4.93 & 4.90 %
    & 0.14 & 0.77 %
    & \textbf{0.04} \\ %
& Yueh Hai Ching Temple Singapore
    & 4.12 & 4.27 %
    & \textbf{0.01} & \textbf{0.01} %
    & 2.44 & 4.28 %
    & \textbf{0.01} & \textbf{0.01} %
    & \textbf{0.01} \\ %
& Average
    & 6.58 & 6.20 %
    & 0.49 & 0.82 %
    & 5.25 & 6.88 %
    & 0.07 & 2.38 %
    & \textbf{0.03} \\ %

\end{tabular}
}

    \caption{Results of Euclidean reconstruction of novel test scenes, with and without data augmentation.
    The results of DPESFM~\cite{Moran_2021_ICCV} have been acquired by us training the model, along with the results reported by~\cite{Moran_2021_ICCV} in parentheses, if available.
    The result of \COLMAP{}, as reported by~\cite{Moran_2021_ICCV}, is also added for reference.}
    \label{tab:learning_euc_inf_ba_with_and_without_aug}
\end{table*}
\subsection{Experimental Setup}
We take a particular interest in Euclidean reconstruction, due to the relevance of
metric reconstructions for %
photogrammetry applications.
Furthermore, we focus mainly on the task of learning and generalizing from a set of training scenes, due to the great potential in having a fast learned model acquire reconstructions of completely novel scenes.
We use the DPESFM method of Moran et~al.~\cite{Moran_2021_ICCV} as a baseline, as they have presented experiments on precisely this.
To this end, we also use Olsson's dataset~\cite{olsson_scia2011} (together with the VGG dataset~\cite{oxford_vgg} for additional projective reconstruction experiments in the supplement), and we use the same random partition of 10 test scenes and 3 validation scenes.

We base our implementation on that of~\cite{Moran_2021_ICCV}, implemented in PyTorch~\cite{pytorch_neurips2019}, but replace the network architecture with our proposed \graphattnsfm{} network, implemented using GATv2Conv layers~\cite{brody2022how} from PyTorch Geometric~\cite{pytorch-geometric}.
The implementation details are in general consistent with~\cite{Moran_2021_ICCV}, for more details see the supplement.
The DPESFM results reproduced by us are obtained by running the method of~\cite{Moran_2021_ICCV} with our code base, with the hyper-parameters published on the official GitHub repository of the method~\cite{dpesfm_github}.
For our model, we use $L=12$ layers and feature dimensions $d_p=32, d_v=1024, d_s=64, d_g=2048$, which corresponds to 145M parameters.
By using small dimensions for the projection features and scene point features the otherwise high memory consumption is significantly reduced.
Hyper-parameters have been chosen based on the performance on the validation data,
but without any rigorous hyper-parameter tuning.

\subsection{Euclidean Reconstruction of Novel Scenes}\label{sec:learning_euc}
\begin{table}[t]
    \centering
    \footnotesize
    
\begin{tabular}{l | rr | rrr}
    Scene & Infer. & BA & \headerCOLMAP{} \\
    Alcatraz Courtyard & 0.24 &   45.54 & 286 \\
    Alcatraz Water Tower & 0.13 &   31.11 & 130 \\
    Drinking Fountain Somewhere In Zurich & 0.06 &    1.98 & 16 \\
    Nijo Castle Gate & 0.09 &    3.97 & 21 \\
    Porta San Donato Bologna & 0.18 &   27.02 & 170 \\
    Round Church Cambridge & 0.43 &   56.47 & 229 \\
    Smolny Cathedral St Petersburg & 0.49 &   86.09 & 516 \\
    Some Cathedral In Barcelona & 0.24 &   47.05 & 451 \\
    Sri Veeramakaliamman Singapore & 0.63 &  115.80 & 583 \\
    Yueh Hai Ching Temple Singapore & 0.08 &    8.54 & 106 \\
\end{tabular}

    \caption{Runtime (s) of our method for Euclidean reconstruction on test scenes, in comparison with \COLMAP{} (measured by~\cite{Moran_2021_ICCV}).}
    \label{tab:exec_times_learning_euc}
\end{table}

In Table~\ref{tab:learning_euc_inf_ba_with_and_without_aug}, we report the Euclidean reconstruction results of our method, when applied on novel test scenes, and measure the performance by reprojection errors, as well as rotation and translation errors for the estimated camera poses.
For calculating the latter, the predicted and ground truth poses are first aligned to a common reference frame, as described in Section~\ref{sec:pose_align} in the supplement.
The results of DPESFM as reported in~\cite{Moran_2021_ICCV} can be seen in parentheses, when available, but we also retrain their model and incorporate further results of the model, not reported in~\cite{Moran_2021_ICCV}, in particular results without bundle adjustment, and additional metrics to reprojection errors.
Experiments are carried out both with and without data augmentation with random rotational cameras perturbations, as described in Section~\ref{sec:data_aug}, and the columns are partitioned accordingly.
Consistently with~\cite{Moran_2021_ICCV}, \emph{Inference} refers to the network output followed by a computationally cheap triangulation and \emph{Inference + BA} refers to the network output followed by bundle adjustment, which is carried out in the same manner as~\cite{Moran_2021_ICCV}, using the Ceres solver~\cite{ceres}.
Moran et al.~\cite{Moran_2021_ICCV} also performed experiments with fine-tuning the network parameters on the test scenes.
In contrast, we advocate against this, since this approach is relatively costly, and one might as well acquire high-quality reconstructions from traditional SfM pipelines such as \COLMAP{}, in similar execution time.
Nevertheless, the supplementary material includes such experiments, for completeness.

Without data augmentation, there is no clear difference between the \emph{Inference} results of both methods.
Neither reconstruction is good, and it is not so meaningful to compare minor differences in the metrics under these circumstances.
In particular, calculating rotation and translation errors requires aligning the predicted and ground truth camera poses in a common reference frame, as described in Section~\ref{sec:pose_align} in the supplement, and when the predicted configuration of camera poses is not very consistent with the ground truth, determining this alignment itself can be sensitive.
In all other scenarios, we outperform DPESFM, and when using both data augmentation and applying bundle adjustment, the solutions we acquire for the novel test scenes are in general very good.
In terms of reprojection error and translation error, we match the performance of \COLMAP{} for almost every test scene, leading to an average reprojection error of 0.85 (vs 0.66) px, and an average translation error of 0.07 (vs 0.03) m.
In terms of rotation error, the error is actually lower than \COLMAP{}'s for 8 out of 10 scenes, but the average error is slightly larger: 0.144 (vs 0.074) degrees.

\subsection{Artificial Outlier Injection}\label{sec:outlier_injection_results}
In Table~\ref{tab:learning_euc_rhaug1520_outliers10} we present additional results where,
on top of data augmentation (see Section~\ref{sec:data_aug}),
we also apply artificial outlier injection during training according to Section~\ref{sec:outlier_injection}.

We evaluate inference of the model in two settings: 1)~Applied on the uncorrupted outlier-free test scenes as-is. 2)~Applied to the test scenes followed by random injection of $10 \%$,
in the same manner as during training\footnote{Yet, during evaluation, we use all true measurements as ground truth targets when calculating reprojection errors, even in the corrupted case.}.
One can immediately observe significantly reduced reprojection errors for our model, suggesting that the outlier injection during training has a regularizing effect.
Beyond that, this experiment is intended to serve as a teaser for the utility and promise of learning-based methods for SfM.
Constant challenges such as the presence of outliers may not be as big of a challenge for learning-based as for conventional methods, and the phenomenon may even be exploited to develop effective training strategies.
In fact, for any real-world scenario, it is crucial for learning-based methods applied on image point correspondences to be resilient to outliers, as we do not have the luxary of facing high-quality curated data such as Olsson's~\cite{olsson_scia2011} in the wild.
Note that it should be possible to apply bundle adjustment on the corrupted test data as well, when combined with a robust loss function.

Similarly as for the data augmentation, we note that DPESFM appears to struggle with the presence of outliers during training, and its performance on the test scenes at inference is not very good.
It is very probable that the nonlinear attention layers make our model more powerful and expressive than DPESFM.
\begin{table}[t]
    \centering
    \footnotesize
    
\begin{tabular}{L{2.45cm} | rr | rr}
& \multicolumn{2}{c|}{Uncorrupted} & \multicolumn{2}{c}{Outlier-injected} \\
& Ours & \headerMORAN{} & Ours & \headerMORAN{} \\
Alcatraz Courtyard & \textbf{47.74} & 85.81 & \textbf{52.99} & 94.24 \\
\cline{1-1}
Alcatraz Water Tower & \textbf{35.96} & 72.84 & \textbf{37.89} & 83.55 \\
\cline{1-1}
Drinking Fountain Somewhere in Zurich & \textbf{52.08} & 1012.14 & \textbf{46.65} & 1453.31 \\
\cline{1-1}
Nijo Castle Gate & \textbf{46.48} & 72.99 & \textbf{62.52} & 126.18 \\
\cline{1-1}
Porta San Donato Bologna & \textbf{53.12} & 88.02 & \textbf{65.08} & 94.72 \\
\cline{1-1}
Round Church Cambridge & \textbf{36.09} & 63.72 & \textbf{48.63} & 90.63 \\
\cline{1-1}
Smolny Cathedral St Petersburg & \textbf{47.28} & 91.03 & \textbf{59.52} & 98.05 \\
\cline{1-1}
Some Cathedral in Barcelona & \textbf{109.86} & 397.75 & \textbf{123.75} & 462.28 \\
\cline{1-1}
Sri Veeramakali-\newline amman Singapore & \textbf{63.60} & 169.63 & \textbf{70.90} & 146.98 \\
\cline{1-1}
Yueh Hai Ching Temple Singapore & \textbf{26.83} & 51.41 & \textbf{36.69} & 57.59 \\
\cline{1-1}
Average & \textbf{51.91} & 210.53 & \textbf{60.46} & 270.75 \\
\end{tabular}

    \caption{Reprojection errors of Euclidean reconstruction of novel test scenes, with model trained with data augmentation as well as artificial outlier injection.
    The results of DPESFM~\cite{Moran_2021_ICCV} have been acquired by us training the model.
    }
    \label{tab:learning_euc_rhaug1520_outliers10}
\end{table}

\subsection{Additional Results}
In the supplementary material we provide additional results on projective reconstruction, as well as single-scene optimization, for completeness.
We also include statistics of the number of views and scene points for each scene.

\section{Conclusion}
\label{sec:conclusion}
With this paper we have added to the research direction of learned initialization-free Structure-from-Motion by introducing a novel and expressive graph attention network outperforming previous learning-based methods for Euclidean reconstruction of novel scenes.
When succeeded by bundle adjustment, we are able to reconstruct novel test scenes to perfection or of very decent precision, at a speedup of about $5-10\times$ compared to \COLMAP{}.
Moreover, we illustrate great potential in coping with the presence of outliers, and even show artificial outlier injection to provide an effective regularizing training strategy, combined with data augmentation by random rotational camera perturbations.

For future work, it could definitely be promising to train on larger datasets, since model generalization is still a challenge.
While the proposed model does not require fine-tuning, and thus is much faster to execute than previous work, it still relies on bundle adjustment to acquire a good reconstruction, which constitutes the computational bottleneck.
Exploring alternative formulations where the network regresses relative camera poses is also an interesting direction, as the current absolute pose prediction may be a weakness due to the reconstruction ambiguity up to a similarity transformation.
Other possible research directions include
unrolling the architecture to multiple learned iterations / refinement steps,
extensions to non-rigid Structure-from-Motion,
and incorporating modern learned image matching pipelines such as \cite{sun2021loftr,bokman2022case,bökman2024steerers} in an end-to-end fashion.

\paragraph{Acknowledgements}
The work was
supported by the Wallenberg AI, Autonomous Systems and Software Program (WASP) funded by the Knut and Alice Wallenberg Foundation,
and the Swedish Research Council grants no. 2016-04445, 2018-05375, and 2023-05341.
Computations were enabled by resources provided by
the National Academic Infrastructure for Supercomputing in Sweden (NAISS)
at Chalmers Centre for Computational Science and Engineering (C3SE) and National Supercomputer Centre (NSC) Berzelius at Link\"oping University
partially funded by the Swedish Research Council through
grant
no. 2022-06725.

\clearpage
{
    \small
    \bibliographystyle{ieeenat_fullname}
    \bibliography{ms}
}

\clearpage
\pagestyle{plain} %
\setcounter{page}{1}
\maketitlesupplementary

\appendix

In this supplementary material, we provide additional implementation details (Section~\ref{sec:impl}), dataset statistics (Section~\ref{sec:data_stats}), and some elaboration on alignment between estimated and ground truth camera poses carried out in certain cases (Section~\ref{sec:pose_align}).
Finally, additional results are presented in  Section~\ref{sec:additional_results}, including visualizations of the reconstructions quantitatively evaluated in the main paper.

\section{Implementation Details}\label{sec:impl}

\subsection{Artificial Outlier Injection Details}
Here we explain the scheme we use when choosing which keypoints to corrupt with artificial outliers:
\begin{enumerate}
    \item Define a desired outlier rate $\eta$, which we set to $10 \%$ in our experiments.
    \item Mark all keypoints in views with $\leq 8$ scene points visible as inliers, and do the same for all projections of scene points visible in $\leq 2$ views. Let $N$ denote the total number of keypoints, and $n_{fixed\_inliers}$ denote the number of keypoints marked as inliers here.
    \item From the $N - n_{fixed\_inliers}$ remaining keypoints, take a random sample of candidates for outlier injection. In order not to violate the view and scene point lower bounds, some of these may not be accepted as candidates, so we add some margin on the number of keypoints selected.
    If selecting
    $n_{sel\_target}$
    out of the remaining $N - n_{fixed\_inliers}$ candidates,
    i.e. a fraction of
    $\nu = \frac{n_{sel\_target}}{N - n_{fixed\_inliers}}$,
    would result in the desired outlier rate $\eta$, we instead sample a fraction
    according to the
    the harmonic average of $\nu$ and $1$, i.e.
    $1 / ( 0.5 \cdot \frac{1}{\nu} + 0.5 \cdot \frac{1}{1} )$..
    \item We now determine whether the lower bounds would be violated for any views or scene points, by marking the candidates as outliers. If so, all projections corresponding to any of those views or scene points are instead fixed to be inliers.
    \item If the remaining number of outlier candidates are still enough to meet the desired outlier rate, a random sample of the candidates is selected, to achieve exactly the desired rate $\eta$. If the number of candidates are not enough, we repeat steps 3.-4. until the condition is met.
\end{enumerate}

\subsection{Training and Learning Rate Schedule}
We train our model for 40k epochs and perform validation every 250 epochs.
Since we are using an attention-based model, we warm up the learning rate for 2500 iterations, linearly from 0 to 1e-4.
Then we apply an exponential learning rate decay corresponding to a factor of 10 every 250k iterations.
In case of fine-tuning, the learning rate is kept constant at 1e-4 through all 1000 iterations.
By the same approach as~\cite{Moran_2021_ICCV}, we randomly sample partial scenes during training, with subsequences of 10-20 views.

\subsection{Computational Resources}
We carry out our experiments on a compute cluster with Nvidia A40 GPUs.

\subsection{Post-Processing}
Post-processing of the network output is done in the same manner as~\cite{Moran_2021_ICCV}.
That is, either a cheap DLT triangulation is carried out, often improving the quality of the reconstruction, or bundle adjustment (BA) is carried out.
BA, if applied, is carried out with Huber loss with threshold 0.1, and in two rounds interleaved with triangulation, according to~\cite{Moran_2021_ICCV}.
This also tends to improve the final results a bit, if global convergence is not achieved during the first round.
Each round of BA is limited to 100 iterations.

\section{Dataset Statistics}\label{sec:data_stats}
In Tables~\ref{tab:scene_stats_euc} and~\ref{tab:scene_stats_proj} we report the number of views and scene points for each of the Euclidean and projective scenes, respectively.
\begin{table}%
    \centering
    \footnotesize
    \begin{tabular}{lrr}
 & \multicolumn{2}{r}{} \\
 & \#Views & \#Scene Points \\
Alcatraz Courtyard & 133 & 23674 \\
Alcatraz Water Tower & 172 & 14828 \\
Buddah Tooth Relic Temple Singapore & 162 & 27920 \\
Doge Palace Venice & 241 & 67107 \\
Door Lund & 12 & 17650 \\
Drinking Fountain Somewhere in Zurich & 14 & 5302 \\
East Indiaman Goteborg & 179 & 25655 \\
Ecole Superior De Guerre & 35 & 13477 \\
Eglise Du Dome & 85 & 84792 \\
Folke Filbyter & 40 & 21150 \\
Fort Channing Gate Singapore & 27 & 23627 \\
Golden Statue Somewhere in Hong Kong & 18 & 39989 \\
Gustav II Adolf & 57 & 5813 \\
Gustav Vasa & 18 & 4249 \\
Jonas Ahlstromer & 40 & 2021 \\
King´s College University of Toronto & 77 & 7087 \\
Lund University Sphinx & 70 & 32668 \\
Nijo Castle Gate & 19 & 7348 \\
Pantheon Paris & 179 & 29383 \\
Park Gate Clermont Ferrand & 34 & 9099 \\
Plaza De Armas Santiago & 240 & 26969 \\
Porta San Donato Bologna & 141 & 25490 \\
Round Church Cambridge & 92 & 84643 \\
Skansen Kronan Gothenburg & 131 & 28371 \\
Smolny Cathedral St Petersburg & 131 & 51115 \\
Some Cathedral in Barcelona & 177 & 30367 \\
Sri Mariamman Singapore & 222 & 56220 \\
Sri Thendayuthapani Singapore & 98 & 88849 \\
Sri Veeramakaliamman Singapore & 157 & 130013 \\
Statue Of Liberty & 134 & 49250 \\
The Pumpkin & 196 & 69341 \\
Thian Hook Keng Temple Singapore & 138 & 34288 \\
Tsar Nikolai I & 98 & 37857 \\
Urban II & 96 & 22284 \\
Vercingetorix & 69 & 10754 \\
Yueh Hai Ching Temple Singapore & 43 & 13774 \\
\end{tabular}
    \caption{Number of views and scene points for the Euclidean scenes.}
    \label{tab:scene_stats_euc}
\end{table}
\begin{table}%
    \centering
    \footnotesize
    \begin{tabular}{lrr}
 & \multicolumn{2}{r}{} \\
 & \#Views & \#Scene Points \\
Alcatraz Courtyard & 133 & 23674 \\
Alcatraz Water Tower & 172 & 14828 \\
Alcatraz West Side Gardens & 419 & 65072 \\
Basilica Di San Petronio & 334 & 46035 \\
Buddah Statue & 322 & 156356 \\
Buddah Tooth Relic Temple Singapore & 162 & 27920 \\
Corridor & 11 & 737 \\
Dinosaur 319 & 36 & 319 \\
Dinosaur 4983 & 36 & 4983 \\
Doge Palace Venice & 241 & 67107 \\
Drinking Fountain Somewhere in Zurich & 14 & 5302 \\
East Indiaman Goteborg & 179 & 25655 \\
Ecole Superior De Guerre & 35 & 13477 \\
Eglise Du Dome & 85 & 84792 \\
Folke Filbyter & 40 & 21150 \\
Golden Statue Somewhere in Hong Kong & 18 & 39989 \\
Gustav II Adolf & 57 & 5813 \\
Gustav Vasa & 18 & 4249 \\
Jonas Ahlstromer & 40 & 2021 \\
King´s College University of Toronto & 77 & 7087 \\
Lund University Sphinx & 70 & 32668 \\
Model House & 10 & 672 \\
Nijo Castle Gate & 19 & 7348 \\
Pantheon Paris & 179 & 29383 \\
Park Gate Clermont Ferrand & 34 & 9099 \\
Plaza De Armas Santiago & 240 & 26969 \\
Porta San Donato Bologna & 141 & 25490 \\
Skansen Kronan Gothenburg & 131 & 28371 \\
Skansen Lejonet Gothenburg & 368 & 74423 \\
Smolny Cathedral St Petersburg & 131 & 51115 \\
Some Cathedral in Barcelona & 177 & 30367 \\
Sri Mariamman Singapore & 222 & 56220 \\
Sri Thendayuthapani Singapore & 98 & 88849 \\
Sri Veeramakaliamman Singapore & 157 & 130013 \\
The Pumpkin & 195 & 69335 \\
Thian Hook Keng Temple Singapore & 138 & 34288 \\
Tsar Nikolai I & 98 & 37857 \\
Urban II & 96 & 22284 \\
\end{tabular}
    \caption{Number of views and scene points for the projective scenes.}
    \label{tab:scene_stats_proj}
\end{table}

\section{Camera Pose Alignment}\label{sec:pose_align}
For both Euclidean and projective reconstruction, there is always a respective inherent ambiguity regarding the choice of coordinate frame.
The loss function that we use is invariant to the choice of coordinate frame, but we also carry out quantitative and qualitative evaluations of the reconstructed scene points and estimated camera poses in the Euclidean setting.
In order to do so, however, one first needs to determine what coordinate frame to use.
To this end, in the same manner as~\cite{Moran_2021_ICCV}, we fit a similarity transformation between estimated and ground truth camera poses, by which the two may be aligned.
It is, however, worth noting that this is not done with a lot of care / robustness, and the alignment may be quite arbitrary in case the pose estimates are inaccurate.

\section{Additional Results}\label{sec:additional_results}
\subsection{Euclidean Reconstruction of Novel Scenes}\label{sec:learning_euc_results_incl_finetune}
In addition to the Euclidean reconstruction results presented in Section~\ref{sec:learning_euc}, which focused on network inference followed by bundle adjustment, in this section we also consider fine-tuning of the network parameters on the novel scenes, before carrying out bundle adjustment.
Fine-tuning is initialized with the model parameters at the epoch of minimal average reprojection error on the validation set.
It should be stressed that this optimization is very costly, and not something that we advocate.
We merely present these experiments to provide a complete comparison with DPESFM~\cite{Moran_2021_ICCV}.
Table~\ref{tab:learning_euc_noaug} presents
reprojection errors, rotation errors and translation errors for models trained without data augmentation,
and
Table~\ref{tab:learning_euc_rhaug1520} presents corresponding
results with data augmentation.
The \emph{Inference} and \emph{Inference BA} results are the same as presented in Section~\ref{sec:learning_euc}, but now presented side-by-side with the results of 1000 iterations of fine-tuning (the \emph{Fine-tune} column), followed by bundle adjustment (column \emph{Fine-tune + BA}).
During fine-tuning, the same loss function is used as during training, but now minimized on the test scene.
For both our method and DPESFM~\cite{Moran_2021_ICCV}, fine-tuning can be used to improve the precision of the reconstruction further, and the results of our model trained with data augmentation and succeeded by bundle adjustment are then on par with \COLMAP{}.
\begin{table*}[t]
    \centering
    \footnotesize
    \resizebox{\textwidth}{!}{%
\begin{tabular}{L{0.13cm}l | rr | rr | rr | rr | r}
 & & \multicolumn{2}{c|}{Inference} & \multicolumn{2}{c|}{Inference + BA} & \multicolumn{2}{c|}{Fine-tune} & \multicolumn{2}{c|}{Fine-tune + BA} &  \\
 
 & & Ours & \headerMORAN{} & Ours & \headerMORAN{} & Ours & \headerMORAN{} & Ours & \headerMORAN{} & \headerCOLMAP{} \\

\hline

\multirow{11}{0.13cm}{\rotatebox{90}{Reprojection error (px)}}
& Alcatraz Courtyard & 68.41 & 68.50 & 1.70 & \textbf{0.81} (0.82) & 2.90 & 5.92 & \textbf{0.81} & \textbf{0.81} (\textbf{0.81}) & \textbf{0.81} \\
& Alcatraz Water Tower & 36.30 & 50.47 & 1.05 & 1.13 (\textbf{0.55}) & 5.83 & 147.83 & \textbf{0.55} & 8.63 (\textbf{0.55}) & \textbf{0.55} \\
& Drinking Fountain Somewhere in Zurich & 36.02 & 45.87 & 0.55 & \textbf{0.31} (7.21) & 0.62 & 27.43 & \textbf{0.31} & 6.73 (\textbf{0.31}) & \textbf{0.31} \\
& Nijo Castle Gate & 65.71 & 64.53 & \textbf{0.73} & \textbf{0.73} (5.81) & 3.82 & 5.22 & \textbf{0.73} & \textbf{0.73} (\textbf{0.73}) & \textbf{0.73} \\
& Porta San Donato Bologna & 74.32 & 94.26 & \textbf{0.74} & \textbf{0.74} (1.10) & 4.42 & 9.37 & \textbf{0.74} & \textbf{0.74} (0.79) & 0.75 \\
& Round Church Cambridge & 61.18 & 55.51 & \textbf{0.39} & \textbf{0.39} (0.50) & 3.92 & 5.92 & \textbf{0.39} & 1.50 (1.51) & \textbf{0.39} \\
& Smolny Cathedral St Petersburg & 156.16 & 120.85 & \textbf{0.81} & \textbf{0.81} (15.15) & 3.33 & 6.68 & \textbf{0.81} & \textbf{0.81} (\textbf{0.81}) & \textbf{0.81} \\
& Some Cathedral in Barcelona & 150.01 & 146.10 & 10.42 & 12.71 (21.46) & 7.99 & 27.80 & \textbf{0.89} & \textbf{0.89} (\textbf{0.89}) & \textbf{0.89} \\
& Sri Veeramakaliamman Singapore & 121.02 & 157.39 & 2.19 & 16.87 (16.92) & 25.06 & 40.76 & 1.20 & 0.88 (17.26) & \textbf{0.71} \\
& Yueh Hai Ching Temple Singapore & 37.46 & 52.59 & \textbf{0.65} & \textbf{0.65} (1.16) & 2.84 & 8.27 & \textbf{0.65} & \textbf{0.65} (\textbf{0.65}) & \textbf{0.65} \\
& Average & 80.66 & 85.61 & 1.92 & 3.52 (7.07) & 6.07 & 28.52 & 0.71 & 2.24 (2.43) & \textbf{0.66} \\

\hline

\multirow{11}{0.13cm}{\rotatebox{90}{Rotation error (deg)}}
& Alcatraz Courtyard
    & 10.102 & 13.201 & 2.607 & \textbf{0.035} & 1.822 & 1.423 & 0.038 & 0.038 & 0.043 \\
& Alcatraz Water Tower
    & 10.637 & 11.053 & 0.499 & 0.764 & 4.161 & 18.217 & 0.228 & 22.764 & \textbf{0.228} \\
& Drinking Fountain Somewhere in Zurich
    & 15.846 & 16.014 & 0.003 & \textbf{0.001} & 0.453 & 19.678 & \textbf{0.001} & 22.776 & 0.007 \\
& Nijo Castle Gate
    & 16.751 & 10.546 & \textbf{0.062} & \textbf{0.062} & 2.301 & 3.326 & 0.064 & 0.064 & 0.064 \\
& Porta San Donato Bologna
    & 23.839 & 24.120 & 0.095 & \textbf{0.094} & 3.664 & 3.410 & \textbf{0.094} & \textbf{0.094} & 0.099 \\
& Round Church Cambridge
    & 18.906 & 14.473 & 0.029 & \textbf{0.026} & 6.720 & 3.116 & 0.028 & 1.089 & 0.035 \\
& Smolny Cathedral St Petersburg
    & 19.387 & 17.971 & 0.023 & \textbf{0.022} & 2.969 & 2.311 & 0.023 & 0.023 & 0.029 \\
& Some Cathedral in Barcelona
    & 27.270 & 30.471 & 10.009 & 20.050 & 3.874 & 18.440 & \textbf{0.019} & \textbf{0.019} & 0.025 \\
& Sri Veeramakaliamman Singapore
    & 28.275 & 36.903 & 0.549 & 4.871 & 15.083 & 30.726 & 0.218 & 0.184 & \textbf{0.169} \\
& Yueh Hai Ching Temple Singapore
    & 15.733 & 22.706 & \textbf{0.038} & \textbf{0.038} & 3.392 & 4.624 & \textbf{0.038} & \textbf{0.038} & 0.043 \\
& Average
    & 18.675 & 19.746 & 1.391 & 2.596 & 4.444 & 10.527 & 0.075 & 4.709 & \textbf{0.074} \\

\hline

\multirow{11}{0.13cm}{\rotatebox{90}{Translation error (m)}}
& Alcatraz Courtyard
    & 4.82 & 4.93 & 1.09 & \textbf{0.01} & 0.49 & 0.48 & \textbf{0.01} & \textbf{0.01} & \textbf{0.01} \\
& Alcatraz Water Tower
    & 8.66 & 7.35 & 0.31 & 0.44 & 2.19 & 8.61 & \textbf{0.12} & 9.60 & \textbf{0.12} \\
& Drinking Fountain Somewhere in Zurich
    & 4.44 & 4.44 & \textbf{0.00} & \textbf{0.00} & 0.04 & 2.58 & \textbf{0.00} & 1.38 & \textbf{0.00} \\
& Nijo Castle Gate
    & 5.07 & 3.08 & \textbf{0.01} & \textbf{0.01} & 0.49 & 0.79 & \textbf{0.01} & \textbf{0.01} & \textbf{0.01} \\
& Porta San Donato Bologna
    & 9.50 & 10.72 & \textbf{0.05} & \textbf{0.05} & 0.70 & 0.83 & \textbf{0.05} & \textbf{0.05} & \textbf{0.05} \\
& Round Church Cambridge
    & 8.94 & 7.19 & \textbf{0.01} & \textbf{0.01} & 1.86 & 1.41 & \textbf{0.01} & 0.56 & \textbf{0.01} \\
& Smolny Cathedral St Petersburg
    & 2.70 & 2.43 & \textbf{0.01} & \textbf{0.01} & 0.14 & 0.24 & \textbf{0.01} & \textbf{0.01} & \textbf{0.01} \\
& Some Cathedral in Barcelona
    & 12.64 & 12.69 & 3.22 & 6.38 & 1.30 & 7.72 & \textbf{0.01} & \textbf{0.01} & \textbf{0.01} \\
& Sri Veeramakaliamman Singapore
    & 4.94 & 4.90 & 0.16 & 1.32 & 3.07 & 4.65 & 0.05 & 0.05 & \textbf{0.04} \\
& Yueh Hai Ching Temple Singapore
    & 4.12 & 4.27 & \textbf{0.01} & \textbf{0.01} & 0.41 & 1.04 & \textbf{0.01} & \textbf{0.01} & \textbf{0.01} \\
& Average
    & 6.58 & 6.20 & 0.49 & 0.82 & 1.07 & 2.84 & \textbf{0.03} & 1.17 & \textbf{0.03} \\

\end{tabular}
}

    \caption{Euclidean reconstruction of novel test scenes, with model trained without data augmentation.
    The results of DPESFM~\cite{Moran_2021_ICCV} have been acquired by us training the model, along with the results reported by~\cite{Moran_2021_ICCV} in parentheses, if available.
    The result of \COLMAP{}, as reported by~\cite{Moran_2021_ICCV}, is also added for reference.
    }
    \label{tab:learning_euc_noaug}
\end{table*}
\begin{table*}[t]
    \centering
    \footnotesize
    \resizebox{\textwidth}{!}{%
\begin{tabular}{L{0.13cm}l | rr | rr | rr | rr | r}
 & & \multicolumn{2}{c|}{Inference} & \multicolumn{2}{c|}{Inference + BA}
 & \multicolumn{2}{c|}{Fine-tune} & \multicolumn{2}{c|}{Fine-tune + BA}
 &  \\

 & & Ours & \headerMORAN{} & Ours & \headerMORAN{} & Ours & \headerMORAN{} & Ours & \headerMORAN{} & \headerCOLMAP{} \\

\hline

\multirow{11}{0.13cm}{\rotatebox{90}{Reprojection error (px)}}
& Alcatraz Courtyard
    & 36.01 & 92.37 %
    & \textbf{0.81} & 0.92 %
    & 2.83 & 4.33 %
    & \textbf{0.81} & \textbf{0.81} %
    & \textbf{0.81} \\ %
& Alcatraz Water Tower
    & 87.67 & 2831.94 %
    & 0.88 & 10.16 %
    & 8.65 & 20.66 %
    & \textbf{0.55} & 0.92 %
    & \textbf{0.55} \\ %
& Drinking Fountain Somewhere in Zurich
    & 219.75 & 234.90 %
    & \textbf{0.31} & 6.73 %
    & 0.91 & 11.40 %
    & \textbf{0.31} & 6.72 %
    & \textbf{0.31} \\ %
& Nijo Castle Gate
    & 61.41 & 68.19 %
    & 0.88 & 0.89 %
    & 3.58 & 4.89 %
    & \textbf{0.73} & \textbf{0.73} %
    & \textbf{0.73} \\ %
& Porta San Donato Bologna
    & 52.15 & 84.46 %
    & 0.76 & 0.75 %
    & 4.53 & 8.01 %
    & \textbf{0.74} & \textbf{0.74} %
    & 0.75 \\ %
& Round Church Cambridge
    & 29.80 & 59.54 %
    & \textbf{0.39} & 1.49 %
    & 3.23 & 5.19 %
    & \textbf{0.39} & 1.54 %
    & \textbf{0.39} \\ %
& Smolny Cathedral St Petersburg
    & 85.38 & 87.81 %
    & \textbf{0.81} & \textbf{0.81} %
    & 2.51 & 3.98 %
    & \textbf{0.81} & \textbf{0.81} %
    & \textbf{0.81} \\ %
& Some Cathedral in Barcelona
    & 125.68 & 687.83 %
    & \textbf{0.89} & 16.77 %
    & 15.73 & 29.11 %
    & \textbf{0.89} & 1.91 %
    & \textbf{0.89} \\ %
& Sri Veeramakaliamman Singapore
    & 83.50 & 166.68 %
    & 2.13 & 9.30 %
    & 23.41 & 43.33 %
    & 0.80 & 4.29 %
    & \textbf{0.71} \\ %
& Yueh Hai Ching Temple Singapore
    & 25.60 & 51.35 %
    & \textbf{0.65} & 0.73 %
    & 3.15 & 8.56 %
    & \textbf{0.65} & \textbf{0.65} %
    & \textbf{0.65} \\ %
& Average
    & 80.69 & 436.51 %
    & 0.85 & 4.86 %
    & 6.85 & 13.95 %
    & 0.67 & 1.91 %
    & \textbf{0.66} \\ %

\hline

\multirow{11}{0.13cm}{\rotatebox{90}{Rotation error (deg)}}
& Alcatraz Courtyard
    & 6.093 & 10.946 & 0.038 & \textbf{0.030} & 2.293 & 1.334 & 0.038 & 0.035 & 0.043 \\
& Alcatraz Water Tower
    & 11.501 & 10.641 & 0.699 & 19.351 & 3.895 & 5.977 & \textbf{0.227} & 0.668 & 0.228 \\
& Drinking Fountain Somewhere in Zurich
    & 15.415 & 15.704 & \textbf{0.001} & 22.759 & 0.488 & 21.769 & \textbf{0.001} & 22.747 & 0.007 \\
& Nijo Castle Gate
    & 17.347 & 20.032 & 0.038 & \textbf{0.036} & 1.036 & 4.537 & 0.064 & 0.063 & 0.064 \\
& Porta San Donato Bologna
    & 18.411 & 25.004 & \textbf{0.094} & \textbf{0.094} & 2.204 & 6.078 & 0.097 & \textbf{0.094} & 0.099 \\
& Round Church Cambridge
    & 10.295 & 18.685 & \textbf{0.029} & 1.086 & 4.827 & 3.794 & 0.030 & 1.158 & 0.035 \\
& Smolny Cathedral St Petersburg
    & 11.662 & 14.380 & 0.023 & \textbf{0.019} & 2.010 & 1.170 & 0.023 & 0.022 & 0.029 \\
& Some Cathedral in Barcelona
    & 27.908 & 29.119 & 0.020 & 47.892 & 16.149 & 24.625 & \textbf{0.019} & 1.762 & 0.025 \\
& Sri Veeramakaliamman Singapore
    & 23.702 & 36.176 & 0.457 & 2.759 & 9.378 & 33.969 & \textbf{0.165} & 1.035 & 0.169 \\
& Yueh Hai Ching Temple Singapore
    & 9.515 & 21.561 & \textbf{0.038} & \textbf{0.038} & 4.103 & 5.770 & \textbf{0.038} & \textbf{0.038} & 0.043 \\
& Average
    & 15.185 & 20.225 & 0.144 & 9.406 & 4.638 & 10.902 & \textbf{0.070} & 2.762 & 0.074 \\

\hline

\multirow{11}{0.13cm}{\rotatebox{90}{Translation error (m)}}
& Alcatraz Courtyard
    & 2.73 & 5.74 & \textbf{0.01} & \textbf{0.01} & 0.60 & 0.42 & \textbf{0.01} & \textbf{0.01} & \textbf{0.01} \\
& Alcatraz Water Tower
    & 7.53 & 7.77 & 0.41 & 9.05 & 2.19 & 3.33 & \textbf{0.12} & 0.40 & \textbf{0.12} \\
& Drinking Fountain Somewhere in Zurich
    & 4.45 & 4.46 & \textbf{0.00} & 1.38 & 0.08 & 1.78 & \textbf{0.00} & 1.38 & \textbf{0.00} \\
& Nijo Castle Gate
    & 5.67 & 6.95 & \textbf{0.01} & \textbf{0.01} & 0.30 & 1.01 & \textbf{0.01} & \textbf{0.01} & \textbf{0.01} \\
& Porta San Donato Bologna
    & 4.80 & 10.48 & \textbf{0.05} & \textbf{0.05} & 0.63 & 1.16 & \textbf{0.05} & \textbf{0.05} & \textbf{0.05} \\
& Round Church Cambridge
    & 5.28 & 9.00 & \textbf{0.01} & 0.56 & 1.47 & 1.48 & \textbf{0.01} & 0.59 & \textbf{0.01} \\
& Smolny Cathedral St Petersburg
    & 2.33 & 2.52 & \textbf{0.01} & \textbf{0.01} & 0.11 & 0.17 & \textbf{0.01} & \textbf{0.01} & \textbf{0.01} \\
& Some Cathedral in Barcelona
    & 12.32 & 12.66 & \textbf{0.01} & 11.93 & 6.67 & 9.82 & \textbf{0.01} & 0.62 & \textbf{0.01} \\
& Sri Veeramakaliamman Singapore
    & 4.93 & 4.90 & 0.14 & 0.77 & 2.28 & 4.83 & \textbf{0.04} & 0.29 & \textbf{0.04} \\
& Yueh Hai Ching Temple Singapore
    & 2.44 & 4.28 & \textbf{0.01} & \textbf{0.01} & 0.49 & 1.32 & \textbf{0.01} & \textbf{0.01} & \textbf{0.01} \\
& Average
    & 5.25 & 6.88 & 0.07 & 2.38 & 1.48 & 2.53 & \textbf{0.03} & 0.34 & \textbf{0.03} \\

\end{tabular}
}

    \caption{Euclidean reconstruction of novel test scenes, with model trained with data augmentation.
    The results of DPESFM~\cite{Moran_2021_ICCV} have been acquired by us training the
    model.
    The result of \COLMAP{}, as reported by~\cite{Moran_2021_ICCV}, is also added for reference.
    }
    \label{tab:learning_euc_rhaug1520}
\end{table*}

\subsection{Projective Reconstruction of Novel Scenes}
In addition to the Euclidean reconstruction results presented in Section~\ref{sec:learning_euc}, in Table~\ref{tab:learning_proj_noaug} we also present corresponding results for projective reconstruction, along with a comparison to DPESFM~\cite{Moran_2021_ICCV} and \VARPRO{}~\cite{VARPRO}.
In Table~\ref{tab:exec_times_learning_proj}, the respective execution times are reported.
In the projective setting we also feed normalized image point correspondences as input to the model but, like~\cite{Moran_2021_ICCV}, use Hartley normalization~\cite{hartley1997} instead of the intrinsic camera parameters.
Note that no data augmentation is applied in the projective setting.
While doing so with random homography transformations would be feasible, determining a distribution over transformations would not be geometrically interpretable in the same way as for the Euclidean setting, and thus less straightforward.
\begin{table*}[t]
    \centering
    \footnotesize
    \begin{tabular}{l | rr | rr | rr | rr | r}

 & \multicolumn{2}{c|}{Inference} & \multicolumn{2}{c|}{Inference + BA} & \multicolumn{2}{c|}{Fine-tune} & \multicolumn{2}{c|}{Fine-tune + BA} &  \\
 & Ours & \headerMORAN{} & Ours & \headerMORAN{} & Ours & \headerMORAN{} & Ours & \headerMORAN{} & \headerVARPRO{} \\
Alcatraz Water Tower & 100.16 & 80.94 & 1.71 & 3.86 (7.37) & 3.48 & 9.83 & \textbf{0.47} & 0.99 (\textbf{0.47}) & \textbf{0.47} \\
Dinosaur 319 & 72.80 & 15.73 & 1.63 & 1.22 (1.58) & 38.98 & 3.84 & 5.88 & 1.32 (1.30) & \textbf{0.43} \\
Dinosaur 4983 & 61.54 & 24.89 & 1.23 & 1.10 (3.99) & 4.83 & 4.49 & 0.64 & 0.92 (1.14) & \textbf{0.42} \\
Drinking Fountain Somewhere in Zurich & 136.73 & 238.87 & \textbf{0.28} & \textbf{0.28} (14.39) & 0.64 & 2.18 & \textbf{0.28} & \textbf{0.28} (\textbf{0.28}) & \textbf{0.28} \\
Eglise Du Dome & 60.29 & 51.72 & 2.49 & 1.15 (2.10) & 4.66 & 7.26 & 0.45 & \textbf{0.29} (1.27) & - \\
Gustav Vasa & 195.08 & 525.96 & 1.63 & 1.62 (6.30) & 1.59 & 3.12 & \textbf{0.16} & \textbf{0.16} (\textbf{0.16}) & \textbf{0.16} \\
Nijo Castle Gate & 98.42 & 80.07 & 0.40 & 0.41 (3.27) & 125.86 & 4.11 & 2.57 & \textbf{0.39} (\textbf{0.39}) & \textbf{0.39} \\
Skansen Kronan Gothenburg & 51.23 & 38.77 & 0.48 & 0.70 (1.64) & 3.42 & 2.74 & \textbf{0.41} & \textbf{0.41} (\textbf{0.41}) & - \\
Some Cathedral in Barcelona & 452.21 & 139.54 & 1.68 & 1.08 (14.87) & 15.35 & 17.23 & 1.51 & 1.67 (\textbf{0.51}) & - \\
Sri Veeramakaliamman Singapore & 247.45 & 165.56 & 4.40 & 5.42 (18.25) & 190.05 & 41.00 & 4.02 & \textbf{3.85} (5.45) & - \\
Average & 147.59 & 136.21 & 1.59 & 1.68 (7.38) & 38.89 & 9.58 & 1.64 & \textbf{1.03} (1.14) & - \\
\end{tabular}

    \caption{Reprojection errors of projective reconstruction of novel test scenes, with model trained without data augmentation.
    The results of DPESFM~\cite{Moran_2021_ICCV} have been acquired by us training the model, along with the results reported by~\cite{Moran_2021_ICCV} in parentheses, if available.
    The result of \VARPRO{}, as reported by~\cite{Moran_2021_ICCV}, is also added for reference.
    }
    \label{tab:learning_proj_noaug}
\end{table*}
\begin{table}[t]
    \centering
    \footnotesize
    
\begin{tabular}{l | rr | rrr}
    Scene & Infer. & BA & \headerGPSFM{} \\
    Alcatraz Water Tower & 0.12 &   84.67 & 137.06 \\
    Dinosaur 319 & 0.07 &    1.43 & 3.25 \\
    Dinosaur 4983 & 0.07 &    6.11 & 4.99 \\
    Eglise Du Dome & 0.42 &   70.91 & 105.84 \\
    Drinking Fountain Somewhere in Zurich & 0.07 &    3.48 & 3.35 \\
    Gustav Vasa & 0.08 &    4.46 & 3.45 \\
    Nijo Castle Gate & 0.13 &    5.89 & 6.37 \\
    Skansen Kronan Gothenburg & 0.27 &   89.31 & 93.83 \\
    Some Cathedral in Barcelona & 0.25 &   91.21 & 110.49 \\
    Sri Veeramakaliamman Singapore & 0.63 &  133.43 & 301.71 \\
    \end{tabular}

    \caption{Runtime (s) of our model for projective reconstruction on test scenes, in comparison with \VARPRO{} (measured by~\cite{Moran_2021_ICCV}).}
    \label{tab:exec_times_learning_proj}
\end{table}
It can be noted that both our method and DPESFM perform worse at inference of test scenes in the projective setting, as compared to the Euclidean one, but for both methods bundle adjustment still converges to relatively good solutions.

\subsection{Single-Scene Recovery}\label{sec:single_scene_optim}
While not of major interest to us due to high computational demand, for completeness we also evaluate our model on single-scene recovery, in line with DPESFM~\cite{Moran_2021_ICCV}.
In this setting, the model is ``trained'' as usual, with the Adam optimizer and the reprojection error loss function with normalized gradients and hinge loss, but on a single scene.
This is indeed very similar to bundle adjustment but with a different parameterization.
Interestingly, however, Moran et al.~\cite{Moran_2021_ICCV} found that even with the direct parameterization with free variables for poses and scene points, these modifications to loss function and optimizer alone can result in much better convergence properties than conventional bundle adjustment, when starting from a random initialization.
To a large extent, this is probably explained by the presence of the hinge loss to overcome the depth barrier in case of scene points with negative depths.

For these experiments, we have used a slightly shallower model, with $L=9$ rather than 12 layers, but the feature dimensions and architecture as a whole remain the same.
The model is optimized for 100k iterations.
Again, warmup is applied by linearly increasing the learning rate from 0 to 1e-4 during the first 2500 iterations, followed by en exponential decay corresponding to a factor of 10 every 35k epochs.
Figure~\ref{tab:optim_euc_reproj} shows the resulting average reprojection errors in pixels, compared both with DPESFM~\cite{Moran_2021_ICCV} as well as other baseline methods reported by~\cite{Moran_2021_ICCV}, i.e. \COLMAP{}~\cite{schoenberger2016sfm,schoenberger2016mvs,colmap_code}, \GESFM{}~\cite{Kasten_2019_ICCV}, and \LINEAR{}~\cite{Jiang_2013_ICCV}.
While in many cases our resulting solution has high precision, there are also quite a few failure cases, which are probably cases of suboptimal local reprojection error minima, which interestingly appears to happen more frequently in the single-scene scenario.
One possibility is that adding more scenes is effectively flattening the loss landscape, although this should be regarded as nothing more than speculation.
In any case, we conclude that our model is more easily trained on multiple scenes simultaneously, in which case we suppose more general geometrical reasoning is encouraged and exploited, and for which our quite expressive model has its edge.
Even if we would achieve better performance, it should be noted that \COLMAP{} works very well, and is much faster to execute than training a model from scratch, albeit not as fast as our learned model combined with bundle adjustment.
In Tables~\ref{tab:optim_euc_roterr} and~\ref{tab:optim_euc_translerr}, the corresponding rotation and translation errors are reported.
Finally, Table~\ref{tab:optim_proj_reproj} shows corresponding results for the projective setting, in which case the results of all baseline methods are again taken as reported by~\cite{Moran_2021_ICCV}.
\begin{table*}[t]
    \centering
    \footnotesize
    \begin{tabular}{l | rrrr | rrrr r}
 & \multicolumn{4}{c|}{Before BA} & \multicolumn{5}{c}{After BA} \\
 & Ours & \headerMORAN{} & \headerGESFM{} & \headerLINEAR{} & Ours & \headerMORAN{} & \headerGESFM{} & \headerLINEAR{} & \headerCOLMAP{} \\

Alcatraz Courtyard & \textbf{0.98} & 1.64 & 66.5 & 16.58 & \textbf{0.81} & \textbf{0.81} & 4.67 & 1.27 & \textbf{0.81} \\
Alcatraz Water Tower & \textbf{1.69} & 2.13 & 131.81 & 56.26 & 0.93 & \textbf{0.55} & 25.93 & 73.72 & \textbf{0.55} \\
Buddah Tooth Relic Temple Singapore & \textbf{1.73} & 2.06 & 89.94 & 47.5 & \textbf{0.85} & \textbf{0.85} & 13.22 & 2.66 & \textbf{0.85} \\
Doge Palace Venice & \textbf{1.28} & 3.62 & 123.53 & - & \textbf{0.98} & 1.00 & 22.32 & - & \textbf{0.98} \\
Door Lund & 11.50 & \textbf{0.32} & (227.0) & 20.89 & 9.94 & \textbf{0.30} & (9.21) & \textbf{0.30} & \textbf{0.30} \\
Drinking Fountain Somewhere in Zurich & 0.36 & \textbf{0.33} & (0.94) & 0.58 & \textbf{0.31} & \textbf{0.31} & (0.27) & \textbf{0.31} & \textbf{0.31} \\
East Indiaman Goteborg & \textbf{1.40} & 4.13 & 170.63 & (94.46) & \textbf{0.89} & 1.85 & 32.37 & (312.9) & \textbf{0.89} \\
Ecole Superior De Guerre & 0.54 & 0.72 & (\textbf{0.35}) & 1.48 & 0.34 & 0.34 & (\textbf{0.14}) & 0.34 & 0.34 \\
Eglise Du Dome & \textbf{0.63} & 0.91 & (90.83) & 26.4 & \textbf{0.27} & \textbf{0.27} & (6.21) & 0.76 & \textbf{0.27} \\
Folke Filbyter & 41.44 & 10.37 & (\textbf{5.74}) & 72.06 & 11.11 & 4.29 & (0.41) & 6.06 & \textbf{0.29} \\
Fort Channing Gate Singapore & \textbf{0.29} & 0.52 & 2.57 & 22.69 & \textbf{0.25} & \textbf{0.25} & \textbf{0.25} & 0.45 & \textbf{0.25} \\
Golden Statue Somewhere in Hong Kong & 0.49 & \textbf{0.40} & 4.98 & 73.7 & \textbf{0.27} & \textbf{0.27} & \textbf{0.27} & 0.3 & \textbf{0.27} \\
Gustav II Adolf & 15.31 & 13.91 & \textbf{6.49} & 31.08 & 11.71 & 11.49 & \textbf{0.26} & \textbf{0.26} & \textbf{0.26} \\
Gustav Vasa & 3.72 & \textbf{3.52} & (5.21) & 11.99 & 3.15 & 3.15 & (\textbf{0.31}) & 0.48 & 0.48 \\
Jonas Ahlstromer & \textbf{10.09} & 10.82 & (36.48) & 236.41 & 7.25 & 8.41 & (0.69) & 4.69 & \textbf{0.22} \\
King´s College University of Toronto & \textbf{0.55} & 0.90 & (11.87) & (27.29) & \textbf{0.34} & \textbf{0.34} & (0.35) & (7.12) & \textbf{0.34} \\
Lund University Sphinx & \textbf{0.75} & 4.78 & 7.19 & 60.64 & \textbf{0.39} & 1.36 & 0.4 & 4.58 & \textbf{0.39} \\
Nijo Castle Gate & \textbf{1.69} & 1.70 & 11.18 & 154.96 & \textbf{0.73} & \textbf{0.73} & \textbf{0.73} & 4.84 & \textbf{0.73} \\
Pantheon Paris & \textbf{0.65} & 1.47 & 79.24 & 39.69 & \textbf{0.49} & \textbf{0.49} & 9.71 & 0.82 & - \\
Park Gate Clermont Ferrand & 7.71 & \textbf{0.57} & 1.71 & 10.5 & 6.75 & \textbf{0.35} & \textbf{0.35} & \textbf{0.35} & \textbf{0.35} \\
Plaza De Armas Santiago & \textbf{7.36} & 7.40 & 146.56 & - & 4.75 & 4.90 & 15.61 & - & 1.13 \\
Porta San Donato Bologna & \textbf{1.12} & 2.28 & 29.5 & 46.12 & \textbf{0.74} & 0.75 & 3.23 & 1.16 & 0.75 \\
Round Church Cambridge & \textbf{2.52} & 2.66 & 19.04 & 9.6 & 1.50 & 1.54 & 2.03 & 0.41 & \textbf{0.39} \\
Skansen Kronan Gothenburg & \textbf{0.74} & 1.24 & 8.82 & (18.49) & \textbf{0.67} & \textbf{0.67} & \textbf{0.67} & (0.69) & \textbf{0.67} \\
Smolny Cathedral St Petersburg & \textbf{0.93} & 1.66 & 19.01 & - & \textbf{0.81} & \textbf{0.81} & 1.0 & - & \textbf{0.81} \\
Some Cathedral in Barcelona & \textbf{1.04} & 2.87 & 47.12 & 66.97 & \textbf{0.89} & \textbf{0.89} & 1.09 & 2.09 & \textbf{0.89} \\
Sri Mariamman Singapore & \textbf{1.36} & 4.13 & 52.13 & 37.16 & \textbf{0.89} & 0.91 & 7.4 & 1.17 & \textbf{0.89} \\
Sri Thendayuthapani Singapore & \textbf{0.87} & 23.37 & (15.93) & 19.57 & 0.67 & 8.44 & (\textbf{0.56}) & 0.72 & 0.67 \\
Sri Veeramakaliamman Singapore & \textbf{2.00} & 3.47 & (205.96) & 18.08 & \textbf{0.71} & 0.73 & (34.72) & 2.2 & \textbf{0.71} \\
Statue Of Liberty & 113.76 & \textbf{26.16} & (1031.8) & 133.81 & 32.51 & 6.97 & (52.05) & 5.08 & \textbf{1.25} \\
The Pumpkin & 13.07 & 33.41 & \textbf{9.71} & (122.54) & 8.67 & 24.85 & \textbf{0.57} & (24.19) & \textbf{0.57} \\
Thian Hook Keng Temple Singapore & 5.18 & \textbf{2.75} & 53.79 & 62.7 & 1.87 & 1.13 & 3.32 & 4.92 & \textbf{1.12} \\
Tsar Nikolai I & 15.40 & 9.79 & \textbf{5.19} & 32.86 & 10.48 & 6.53 & \textbf{0.33} & \textbf{0.33} & \textbf{0.33} \\
Urban II & 17.58 & \textbf{9.38} & 31.71 & 176.19 & 12.70 & 6.92 & 0.72 & 17.61 & \textbf{0.38} \\
Vercingetorix & 6.96 & \textbf{5.08} & 15.87 & 65.57 & 5.02 & 1.50 & 0.54 & 2.93 & \textbf{0.23} \\
Yueh Hai Ching Temple Singapore & \textbf{0.87} & 0.94 & (27.32) & 45.19 & \textbf{0.65} & \textbf{0.65} & (1.64) & 2.06 & \textbf{0.65} \\
\end{tabular}
    \caption{Results of single-scene recovery of Euclidean scenes (average reprojection errors in pixels), compared with DPESFM~\cite{Moran_2021_ICCV}, \GESFM{}~\cite{Kasten_2019_ICCV}, \LINEAR{}~\cite{Jiang_2013_ICCV}, and \COLMAP{}~\cite{schoenberger2016sfm,schoenberger2016mvs,colmap_code}.
    Parentheses mark scenes for which a baseline method has disregarded at least $10\%$ of the cameras.}
    \label{tab:optim_euc_reproj}
\end{table*}
\begin{table*}[t]
    \centering
    \footnotesize
    \begin{tabular}{l | rrrr | rrrr r}
 & \multicolumn{4}{c|}{Before BA} & \multicolumn{5}{c}{After BA} \\

 & Ours & \headerMORAN{} & \headerGESFM{} & \headerLINEAR{} & Ours & \headerMORAN{} & \headerGESFM{} & \headerLINEAR{} & \headerCOLMAP{} \\
Alcatraz Courtyard & \textbf{0.424} & 0.619 & 1.851 & 0.729 & 0.038 & 0.049 & 0.533 & \textbf{0.042} & 0.043 \\
Alcatraz Water Tower & 1.763 & \textbf{0.933} & 1.136 & 1.525 & 0.677 & 0.230 & 9.997 & 1.525 & \textbf{0.228} \\
Buddah Tooth Relic Temple Singapore & 1.545 & \textbf{1.030} & 2.95 & 2.058 & \textbf{0.081} & \textbf{0.081} & 4.709 & 0.551 & 0.083 \\
Doge Palace Venice & \textbf{0.345} & 1.163 & 2.75 & - & 0.048 & 0.211 & 5.317 & - & \textbf{0.031} \\
Door Lund & 13.940 & \textbf{0.024} & (2.041) & 1.148 & 12.844 & 0.006 & (7.552) & \textbf{0.005} & \textbf{0.005} \\
Drinking Fountain Somewhere in Zurich & 0.119 & \textbf{0.031} & (0.054) & 0.077 & \textbf{0.001} & 0.007 & (0.01) & 0.007 & 0.007 \\
East Indiaman Goteborg & \textbf{1.426} & 3.814 & 11.129 & (3.284) & \textbf{0.251} & 3.117 & 12.396 & (3.284) & \textbf{0.251} \\
Ecole Superior De Guerre & 0.243 & 0.318 & (\textbf{0.057}) & 0.182 & \textbf{0.018} & 0.024 & (0.035) & 0.024 & 0.024 \\
Eglise Du Dome & \textbf{0.801} & 0.808 & (2.851) & 0.903 & \textbf{0.031} & 0.037 & (3.631) & 0.162 & 0.036 \\
Folke Filbyter & 74.307 & 74.596 & (\textbf{0.332}) & 1.94 & 68.096 & 70.157 & (0.148) & 4.484 & \textbf{0.036} \\
Fort Channing Gate Singapore & \textbf{0.063} & 0.207 & 0.295 & 0.659 & 0.010 & \textbf{0.020} & \textbf{0.02} & 0.029 & \textbf{0.020} \\
Golden Statue Somewhere in Hong Kong & 0.692 & \textbf{0.292} & 0.669 & 8.264 & 0.024 & 0.031 & 0.03 & \textbf{0.022} & 0.031 \\
Gustav II Adolf & 88.777 & 67.784 & \textbf{0.435} & 1.398 & 66.266 & 58.458 & \textbf{0.021} & \textbf{0.021} & \textbf{0.021} \\
Gustav Vasa & 39.767 & 34.181 & (\textbf{0.841}) & 1.658 & 32.316 & 32.266 & (\textbf{0.751}) & 0.839 & 0.841 \\
Jonas Ahlstromer & 44.994 & 50.190 & (\textbf{1.994}) & 10.154 & 49.640 & 47.117 & (0.082) & 5.391 & \textbf{0.036} \\
King´s College University of Toronto & 1.097 & 0.989 & (\textbf{0.645}) & (1.07) & 0.083 & 0.085 & (\textbf{0.059}) & (4.624) & 0.084 \\
Lund University Sphinx & 0.806 & 19.522 & \textbf{0.738} & 3.476 & \textbf{0.025} & 8.752 & 0.058 & 5.452 & 0.033 \\
Nijo Castle Gate & 0.959 & 1.495 & \textbf{0.399} & 2.097 & \textbf{0.064} & 0.069 & \textbf{0.064} & 0.744 & \textbf{0.064} \\
Pantheon Paris & 0.334 & \textbf{0.192} & 3.766 & 2.655 & \textbf{0.038} & 0.040 & 3.208 & 0.072 & - \\
Park Gate Clermont Ferrand & 25.295 & 0.391 & \textbf{0.203} & 0.296 & 24.779 & \textbf{0.049} & \textbf{0.049} & \textbf{0.049} & \textbf{0.049} \\
Plaza De Armas Santiago & \textbf{5.964} & 6.782 & 6.291 & - & 2.299 & 2.556 & 6.344 & - & \textbf{0.122} \\
Porta San Donato Bologna & \textbf{0.377} & 2.153 & 1.013 & 1.381 & \textbf{0.093} & 0.095 & 0.513 & 0.149 & 0.099 \\
Round Church Cambridge & 2.182 & 2.451 & 1.021 & \textbf{0.634} & 1.099 & 1.107 & 1.851 & \textbf{0.033} & 0.035 \\
Skansen Kronan Gothenburg & \textbf{0.235} & 0.736 & 0.549 & (0.679) & \textbf{0.017} & 0.026 & 0.025 & (0.02) & 0.025 \\
Smolny Cathedral St Petersburg & \textbf{0.469} & 0.554 & 0.493 & - & \textbf{0.023} & 0.033 & 0.028 & - & 0.029 \\
Some Cathedral in Barcelona & \textbf{0.180} & 0.880 & 1.519 & 3.126 & \textbf{0.019} & 0.026 & 0.031 & 0.057 & 0.025 \\
Sri Mariamman Singapore & \textbf{1.003} & 2.302 & 1.433 & 1.615 & \textbf{0.075} & 0.077 & 2.158 & 0.083 & 0.078 \\
Sri Thendayuthapani Singapore & \textbf{0.835} & 46.269 & (1.561) & 1.581 & \textbf{0.137} & 44.170 & (0.329) & 0.138 & 0.138 \\
Sri Veeramakaliamman Singapore & 1.876 & 2.559 & (1.807) & \textbf{0.519} & \textbf{0.167} & 0.175 & (3.41) & 0.288 & 0.169 \\
Statue Of Liberty & 75.495 & 46.887 & (3.449) & \textbf{3.357} & 73.142 & 9.091 & (8.281) & 2.945 & \textbf{0.213} \\
The Pumpkin & 12.650 & 94.672 & \textbf{2.036} & (4.215) & 9.136 & 98.862 & 0.092 & (3.123) & \textbf{0.091} \\
Thian Hook Keng Temple Singapore & 3.691 & \textbf{0.832} & 2.751 & 3.047 & 0.386 & \textbf{0.081} & 0.245 & 0.424 & 0.084 \\
Tsar Nikolai I & 72.322 & 48.499 & \textbf{0.475} & 1.437 & 74.349 & 36.280 & \textbf{0.018} & \textbf{0.018} & \textbf{0.018} \\
Urban II & 60.201 & 47.490 & \textbf{2.077} & 8.951 & 59.713 & 48.214 & 0.175 & 16.348 & \textbf{0.107} \\
Vercingetorix & 91.624 & 69.328 & \textbf{2.203} & 2.365 & 82.565 & 17.706 & 1.431 & 7.138 & \textbf{0.048} \\
Yueh Hai Ching Temple Singapore & 0.752 & \textbf{0.720} & (1.813) & 1.92 & \textbf{0.038} & 0.043 & (0.075) & 0.26 & 0.043 \\
\end{tabular}

    \caption{Results of single-scene recovery of Euclidean scenes (rotation errors in degrees), compared with DPESFM~\cite{Moran_2021_ICCV}, \GESFM{}~\cite{Kasten_2019_ICCV}, \LINEAR{}~\cite{Jiang_2013_ICCV}, and \COLMAP{}~\cite{schoenberger2016sfm,schoenberger2016mvs,colmap_code}.
    Parentheses mark scenes for which a baseline method has disregarded at least $10\%$ of the cameras.}
    \label{tab:optim_euc_roterr}
\end{table*}
\begin{table*}[t]
    \centering
    \footnotesize
    \begin{tabular}{l | rrrr | rrrr r}
 & \multicolumn{4}{c|}{Before BA} & \multicolumn{5}{c}{After BA} \\

 & Ours & \headerMORAN{} & \headerGESFM{} & \headerLINEAR{} & Ours & \headerMORAN{} & \headerGESFM{} & \headerLINEAR{} & \headerCOLMAP{} \\
Alcatraz Courtyard & \textbf{0.103} & 0.160 & 0.767 & 0.378 & \textbf{0.014} & 0.015 & 0.259 & \textbf{0.014} & \textbf{0.014} \\
Alcatraz Water Tower & 1.088 & \textbf{0.518} & 8.332 & 1.643 & 0.393 & 0.116 & 9.147 & 1.643 & \textbf{0.115} \\
Buddah Tooth Relic Temple Singapore & 0.313 & \textbf{0.233} & 2.124 & 1.325 & 0.015 & \textbf{0.014} & 1.429 & 0.125 & 0.015 \\
Doge Palace Venice & \textbf{0.084} & 0.342 & 1.688 & - & 0.014 & 0.029 & 1.608 & - & \textbf{0.012} \\
Door Lund & 1.009 & \textbf{0.006} & (1.603) & 0.226 & 1.386 & \textbf{0.001} & (0.973) & \textbf{0.001} & \textbf{0.001} \\
Drinking Fountain Somewhere in Zurich & 0.012 & \textbf{0.004} & (0.016) & 0.024 & \textbf{0.002} & \textbf{0.002} & (\textbf{0.002}) & \textbf{0.002} & \textbf{0.002} \\
East Indiaman Goteborg & \textbf{0.348} & 0.621 & 2.783 & (2.235) & \textbf{0.065} & 0.509 & 3.099 & (2.235) & \textbf{0.065} \\
Ecole Superior De Guerre & 0.066 & 0.081 & (\textbf{0.006}) & 0.048 & 0.005 & 0.005 & (\textbf{0.002}) & 0.005 & 0.005 \\
Eglise Du Dome & 0.212 & 0.205 & (1.958) & \textbf{0.128} & \textbf{0.010} & \textbf{0.010} & (1.425) & 0.046 & \textbf{0.010} \\
Folke Filbyter & 0.123 & 0.125 & (\textbf{0.003}) & 0.021 & 0.110 & 0.118 & (\textbf{0.000}) & 0.123 & \textbf{0.000} \\
Fort Channing Gate Singapore & \textbf{0.027} & 0.093 & 0.092 & 0.139 & \textbf{0.00008} & \textbf{0.008} & \textbf{0.008} & 0.013 & \textbf{0.008} \\
Golden Statue Somewhere in Hong Kong & 0.120 & \textbf{0.073} & 0.118 & 1.153 & \textbf{0.004} & \textbf{0.004} & \textbf{0.004} & \textbf{0.004} & \textbf{0.004} \\
Gustav II Adolf & 13.324 & 9.714 & \textbf{0.134} & 0.333 & 9.717 & 8.524 & \textbf{0.004} & \textbf{0.004} & \textbf{0.004} \\
Gustav Vasa & 1.235 & 1.085 & (\textbf{0.079}) & 0.266 & 1.136 & 1.145 & (0.101) & \textbf{0.099} & 0.1 \\
Jonas Ahlstromer & 8.649 & 10.888 & (\textbf{0.35}) & 0.895 & 10.405 & 10.451 & (0.01) & 1.259 & 0.011 \\
King´s College University of Toronto & 0.159 & 0.235 & (\textbf{0.152}) & (1.781) & \textbf{0.017} & \textbf{0.017} & (\textbf{0.005}) & (1.877) & \textbf{0.017} \\
Lund University Sphinx & \textbf{0.218} & 4.585 & 0.228 & 1.199 & \textbf{0.009} & 2.191 & 0.016 & 1.512 & \textbf{0.009} \\
Nijo Castle Gate & 0.195 & 0.286 & \textbf{0.141} & 0.348 & \textbf{0.011} & 0.012 & \textbf{0.011} & 0.19 & \textbf{0.011} \\
Pantheon Paris & \textbf{0.029} & 0.050 & 0.867 & 1.275 & \textbf{0.005} & \textbf{0.005} & 0.595 & 0.011 & - \\
Park Gate Clermont Ferrand & 11.706 & 0.125 & \textbf{0.083} & 0.1 & 11.391 & \textbf{0.022} & \textbf{0.022} & \textbf{0.022} & \textbf{0.022} \\
Plaza De Armas Santiago & 2.634 & 2.944 & \textbf{2.45} & - & 1.252 & 1.383 & 2.244 & - & \textbf{0.048} \\
Porta San Donato Bologna & \textbf{0.097} & 0.388 & 0.949 & 1.588 & \textbf{0.046} & \textbf{0.046} & 0.169 & 0.067 & 0.047 \\
Round Church Cambridge & 0.926 & 1.003 & 0.486 & \textbf{0.217} & 0.570 & 0.582 & 0.493 & \textbf{0.012} & \textbf{0.012} \\
Skansen Kronan Gothenburg & \textbf{0.071} & 0.226 & 0.223 & (0.234) & 0.008 & 0.008 & 0.008 & (\textbf{0.007}) & 0.008 \\
Smolny Cathedral St Petersburg & \textbf{0.021} & 0.051 & 0.209 & - & \textbf{0.006} & \textbf{0.006} & 0.007 & - & \textbf{0.006} \\
Some Cathedral in Barcelona & \textbf{0.063} & 0.315 & 1.776 & 1.261 & 0.011 & 0.011 & \textbf{0.013} & 0.024 & 0.010 \\
Sri Mariamman Singapore & \textbf{0.244} & 0.683 & 1.758 & 0.721 & \textbf{0.023} & \textbf{0.023} & 0.614 & 0.025 & \textbf{0.023} \\
Sri Thendayuthapani Singapore & \textbf{0.154} & 3.812 & (0.285) & 0.375 & \textbf{0.034} & 2.870 & (0.053) & \textbf{0.034} & \textbf{0.034} \\
Sri Veeramakaliamman Singapore & 0.432 & 0.597 & (1.966) & \textbf{0.273} & \textbf{0.038} & 0.040 & (1.388) & 0.095 & \textbf{0.038} \\
Statue Of Liberty & 27.130 & 20.012 & (4.55) & \textbf{3.031} & 28.350 & 4.122 & (4.782) & 28.049 & \textbf{0.099} \\
The Pumpkin & 2.881 & 14.890 & \textbf{0.513} & (1.656) & 2.223 & 14.952 & \textbf{0.022} & (14.862) & \textbf{0.022} \\
Thian Hook Keng Temple Singapore & 0.450 & \textbf{0.082} & 0.519 & 0.404 & 0.055 & \textbf{0.008} & 0.024 & 0.043 & \textbf{0.008} \\
Tsar Nikolai I & 12.843 & 9.467 & \textbf{0.219} & 0.261 & 13.990 & 7.836 & \textbf{0.005} & \textbf{0.005} & \textbf{0.005} \\
Urban II & 11.350 & 9.467 & \textbf{0.774} & 2.044 & 10.882 & 9.586 & 0.036 & 3.038 & \textbf{0.021} \\
Vercingetorix & 11.202 & 8.788 & \textbf{1.158} & 2.786 & 9.696 & 3.104 & 0.3 & 1.564 & \textbf{0.011} \\
Yueh Hai Ching Temple Singapore & 0.099 & \textbf{0.098} & (0.642) & 0.303 & \textbf{0.014} & \textbf{0.014} & (0.023) & 0.059 & \textbf{0.014} \\
\end{tabular}

    \caption{Results of single-scene recovery of Euclidean scenes (translation errors in meters), compared with DPESFM~\cite{Moran_2021_ICCV}, \GESFM{}~\cite{Kasten_2019_ICCV}, \LINEAR{}~\cite{Jiang_2013_ICCV}, and \COLMAP{}~\cite{schoenberger2016sfm,schoenberger2016mvs,colmap_code}.
    Parentheses mark scenes for which a baseline method has disregarded at least $10\%$ of the cameras.}
    \label{tab:optim_euc_translerr}
\end{table*}

\begin{table*}[t]
    \centering
    \footnotesize
    \begin{tabular}{l | rrr | rrrrr r}
 & \multicolumn{3}{c|}{Before BA} & \multicolumn{5}{c}{After BA} \\
 & Ours & \headerMORAN{} & \headerGPSFM{} & Ours & \headerMORAN{} & \headerGPSFM{} & \headerPPSFM{} & \headerVARPRO{} \\

Alcatraz Courtyard & \textbf{0.77} & 1.55 & 20.34 & \textbf{0.51} & 0.52 & 0.52 & 0.57 & 0.52 \\
Alcatraz Water Tower & \textbf{1.00} & 2.18 & 16.50 & \textbf{0.47} & \textbf{0.47} & 0.63 & 0.59 & \textbf{0.47} \\
Alcatraz West Side Gardens & \textbf{1.27} & 9.54 & 1007.50 & \textbf{0.72} & 0.76 & 326.99 & 1.77 & - \\
Basilica Di San Petronio & \textbf{6.78} & 7.90 & 1871.41 & 1.14 & 0.96 & 60.69 & \textbf{0.63} & - \\
Buddah Statue & 418.74 & \textbf{18.88} & 919.26 & 7.73 & 2.93 & 96.96 & \textbf{0.41} & - \\
Buddah Tooth Relic Temple Singapore & \textbf{1.77} & 4.59 & 18.53 & \textbf{0.59} & 0.60 & 0.62 & 0.71 & 0.60 \\
Corridor & 0.32 & \textbf{0.30} & 0.64 & \textbf{0.26} & \textbf{0.26} & \textbf{0.26} & 0.27 & \textbf{0.26} \\
Dinosaur 319 & \textbf{1.25} & 2.35 & 4.66 & 0.93 & 1.53 & \textbf{0.43} & 0.47 & \textbf{0.43} \\
Dinosaur 4983 & 4.99 & 1.96 & \textbf{1.54} & 0.95 & 0.57 & \textbf{0.42} & 0.47 & \textbf{0.42} \\
Doge Palace Venice & \textbf{1.45} & 3.60 & 170.93 & \textbf{0.60} & \textbf{0.60} & 3.52 & 0.67 & - \\
Drinking Fountain Somewhere in Zurich & \textbf{0.32} & 0.33 & 1.29 & \textbf{0.28} & \textbf{0.28} & \textbf{0.28} & 0.31 & \textbf{0.28} \\
East Indiaman Goteborg & 27.44 & \textbf{3.31} & 99.38 & 3.23 & 0.99 & 5.11 & \textbf{0.67} & - \\
Ecole Superior De Guerre & \textbf{0.56} & 0.75 & 1.88 & \textbf{0.26} & \textbf{0.26} & \textbf{0.26} & 0.28 & \textbf{0.26} \\
Eglise Du Dome & \textbf{0.95} & 1.10 & 8.41 & \textbf{0.23} & 0.24 & 0.24 & 0.25 & - \\
Folke Filbyter & 115.59 & 8.87 & \textbf{1.78} & 8.76 & 8.58 & 0.82 & \textbf{0.33} & 277.89 \\
Golden Statue Somewhere in Hong Kong & 0.56 & \textbf{0.35} & 0.81 & \textbf{0.22} & \textbf{0.22} & \textbf{0.22} & 0.24 & \textbf{0.22} \\
Gustav II Adolf & 196.68 & 14.77 & \textbf{5.91} & 8.10 & 5.83 & \textbf{0.23} & 0.24 & \textbf{0.23} \\
Gustav Vasa & 0.56 & \textbf{0.23} & 1.82 & \textbf{0.16} & \textbf{0.16} & \textbf{0.16} & 0.17 & \textbf{0.16} \\
Jonas Ahlstromer & 16.16 & \textbf{14.38} & 28.83 & 5.37 & 4.72 & \textbf{0.18} & 0.20 & \textbf{0.18} \\
King´s College University of Toronto & \textbf{2.19} & 2.27 & 22.89 & 0.51 & 0.78 & 2.35 & 0.26 & \textbf{0.24} \\
Lund University Sphinx & 81.34 & \textbf{3.64} & 10.00 & 0.82 & \textbf{0.34} & 0.45 & 0.37 & \textbf{0.34} \\
Model House & 0.49 & \textbf{0.37} & 3.66 & \textbf{0.34} & \textbf{0.34} & 1.12 & 0.40 & \textbf{0.34} \\
Nijo Castle Gate & 1.21 & \textbf{0.71} & 20.08 & \textbf{0.39} & \textbf{0.39} & \textbf{0.39} & 0.43 & \textbf{0.39} \\
Pantheon Paris & \textbf{0.96} & 1.75 & 44.85 & \textbf{0.48} & 0.49 & 2.85 & 0.62 & - \\
Park Gate Clermont Ferrand & \textbf{0.47} & 0.61 & 13.82 & \textbf{0.31} & \textbf{0.31} & 0.32 & 0.49 & \textbf{0.31} \\
Plaza De Armas Santiago & \textbf{2.42} & 5.10 & 81.01 & 0.69 & \textbf{0.64} & 3.14 & 0.71 & - \\
Porta San Donato Bologna & \textbf{0.90} & 1.58 & 33.36 & \textbf{0.40} & \textbf{0.40} & 0.61 & 3.75 & \textbf{0.40} \\
Skansen Kronan Gothenburg & \textbf{0.59} & 1.19 & 8.90 & \textbf{0.41} & 0.41 & 0.44 & 0.44 & - \\
Skansen Lejonet Gothenburg & 8.22 & 10.82 & 69.81 & \textbf{1.21} & 2.05 & 7.48 & 1.28 & - \\
Smolny Cathedral St Petersburg & \textbf{0.55} & 1.66 & 83.78 & \textbf{0.46} & \textbf{0.46} & \textbf{0.46} & 0.50 & - \\
Some Cathedral in Barcelona & \textbf{0.82} & 3.67 & 14.77 & \textbf{0.51} & \textbf{0.51} & \textbf{0.51} & 0.54 & - \\
Sri Mariamman Singapore & \textbf{2.39} & 7.06 & 39.89 & 1.02 & \textbf{0.61} & 0.78 & 0.85 & - \\
Sri Thendayuthapani Singapore & 3.30 & \textbf{2.12} & 13.25 & 2.08 & \textbf{0.31} & 0.56 & 0.33 & - \\
Sri Veeramakaliamman Singapore & 13.80 & \textbf{6.47} & 99.99 & 3.98 & \textbf{0.52} & 1.78 & 0.66 & - \\
The Pumpkin & 3516.42 & 14.45 & \textbf{8.97} & 75.77 & \textbf{0.38} & \textbf{0.38} & 0.42 & - \\
Thian Hook Keng Temple Singapore & 7.99 & \textbf{7.59} & 26.78 & \textbf{0.54} & \textbf{0.54} & 0.55 & 0.66 & \textbf{0.54} \\
Tsar Nikolai I & 25.75 & \textbf{6.04} & 13.21 & 4.03 & 2.43 & 0.33 & 0.31 & \textbf{0.29} \\
Urban II & 487.12 & \textbf{16.91} & 87.25 & 23.68 & 6.84 & \textbf{0.27} & 0.31 & 3.61 \\
\end{tabular}

    \caption{Results of single-scene recovery of projective scenes (average reprojection errors in pixels), compared with DPESFM~\cite{Moran_2021_ICCV}, \GPSFM{}~\cite{Kasten_2019_CVPR}, and \PPSFM{}~\cite{magerand2017}, and \VARPRO{}~\cite{VARPRO}.}
    \label{tab:optim_proj_reproj}
\end{table*}

\subsection{Visualizations of Novel Scene Reconstructions}\label{sec:learning_euc_viz}
In Figures~{\ref{fig:AlcatrazCourtyard}-\ref{fig:YuehHaiChingTempleSingapore}}, we provide visualizations for the Euclidean reconstructions of all 10 novel test scenes, corresponding to the results reported in Sections~\ref{sec:learning_euc} and\ref{sec:outlier_injection_results}, both for our model as well as DPESFM~\cite{Moran_2021_ICCV}.
All point clouds are predictions, together with estimated (red) and ground truth (blue) camera poses.
Some outlier filtering has been carried out on the plotted point clouds by coordinate-wise quantiles.
Again, please note that the ground truth poses are aligned with the predicted cameras according to Section~\ref{sec:pose_align}, and as such their relative motion is only meaningful if the predicted poses are good enough for the alignment to be accurately estimated (typically after BA but not always before).
It can be observed from the plots that our method combined with BA can typically recover high quality scene structures and camera poses, especially when combined with data augmentation.
While DPESFM~\cite{Moran_2021_ICCV} combined with BA also often works relatively well, one notable failure case is \emph{Sri Veeramakaliamman Singapore}, which is a very large scene, that we however manage to recover quite descently.
Also note that while our results are in general improved by incorporating data augmentation, it is quite noticeable that DPESFM breaks down from this, possibly due to a limited model capacity / expressivity.
In particular, with the help of data augmentation, we are able to recover the \emph{Some Cathedral in Barcelona} scene very well, while DPESFM still fails, and in general shows deterioated results whenever data augmentation is applied.
\begin{figure*}
    \centering
    \renewcommand{\sceneid}{AlcatrazCourtyard}
    \renewcommand{\scenename}{Alcatraz Courtyard}
    \renewcommand{\trimleft}{8}
    \renewcommand{\trimbottom}{8}
    \renewcommand{\trimright}{8}
    \renewcommand{\trimtop}{8}
    \input{figures/viz/all_in_one.tex}
    \caption{Euclidean reconstruction of \emph{\scenename{}}. \ref{fig:\sceneid_ours_inf_noaug}-\ref{fig:\sceneid_ours_inf_aug} and \ref{fig:\sceneid_ours_inf_noaug}-\ref{fig:\sceneid_ours_inf_aug} show the results of our method with and without data augmentation and bundle adjustment, while \ref{fig:\sceneid_ours_inf_out_uncorrupt} and \ref{fig:\sceneid_ours_inf_out_corrupt} show the results of training with artificially injected outliers, with / without corruption of the test scene as well. Finally, the corresponding results of DPESFM~\cite{Moran_2021_ICCV} can be seen in \ref{fig:\sceneid_baseline_inf_noaug}-\ref{fig:\sceneid_baseline_inf_out_corrupt}.}
    \label{fig:\sceneid}
\end{figure*}
\begin{figure*}
    \centering
    \renewcommand{\sceneid}{AlcatrazWaterTower}
    \renewcommand{\scenename}{Alcatraz Water Tower}
    \renewcommand{\trimleft}{6}
    \renewcommand{\trimright}{6}
    \renewcommand{\trimleft}{5.5}
    \renewcommand{\trimbottom}{5.5}
    \renewcommand{\trimright}{5.5}
    \renewcommand{\trimtop}{5.5}
    \input{figures/viz/all_in_one.tex}
    \caption{Euclidean reconstruction of \emph{\scenename{}}. \ref{fig:\sceneid_ours_inf_noaug}-\ref{fig:\sceneid_ours_inf_aug} and \ref{fig:\sceneid_ours_inf_noaug}-\ref{fig:\sceneid_ours_inf_aug} show the results of our method with and without data augmentation and bundle adjustment, while \ref{fig:\sceneid_ours_inf_out_uncorrupt} and \ref{fig:\sceneid_ours_inf_out_corrupt} show the results of training with artificially injected outliers, with / without corruption of the test scene as well. Finally, the corresponding results of DPESFM~\cite{Moran_2021_ICCV} can be seen in \ref{fig:\sceneid_baseline_inf_noaug}-\ref{fig:\sceneid_baseline_inf_out_corrupt}.}
    \label{fig:\sceneid}
\end{figure*}
\begin{figure*}
    \centering
    \renewcommand{\sceneid}{DrinkingFountainSomewhereInZurich}
    \renewcommand{\scenename}{Drinking Fountain Somewhere in Zurich}
    \renewcommand{\trimleft}{6}
    \renewcommand{\trimright}{6}
    \renewcommand{\trimleft}{6}
    \renewcommand{\trimbottom}{6}
    \renewcommand{\trimright}{6}
    \renewcommand{\trimtop}{6}
    \input{figures/viz/all_in_one.tex}
    \caption{Euclidean reconstruction of \emph{\scenename{}}. \ref{fig:\sceneid_ours_inf_noaug}-\ref{fig:\sceneid_ours_inf_aug} and \ref{fig:\sceneid_ours_inf_noaug}-\ref{fig:\sceneid_ours_inf_aug} show the results of our method with and without data augmentation and bundle adjustment, while \ref{fig:\sceneid_ours_inf_out_uncorrupt} and \ref{fig:\sceneid_ours_inf_out_corrupt} show the results of training with artificially injected outliers, with / without corruption of the test scene as well. Finally, the corresponding results of DPESFM~\cite{Moran_2021_ICCV} can be seen in \ref{fig:\sceneid_baseline_inf_noaug}-\ref{fig:\sceneid_baseline_inf_out_corrupt}.}
    \label{fig:\sceneid}
\end{figure*}
\begin{figure*}
    \centering
    \renewcommand{\sceneid}{NijoCastleGate}
    \renewcommand{\scenename}{Nijo Castle Gate}
    \renewcommand{\trimleft}{9}
    \renewcommand{\trimbottom}{9}
    \renewcommand{\trimright}{9}
    \renewcommand{\trimtop}{9}
    \input{figures/viz/all_in_one.tex}
    \caption{Euclidean reconstruction of \emph{\scenename{}}. \ref{fig:\sceneid_ours_inf_noaug}-\ref{fig:\sceneid_ours_inf_aug} and \ref{fig:\sceneid_ours_inf_noaug}-\ref{fig:\sceneid_ours_inf_aug} show the results of our method with and without data augmentation and bundle adjustment, while \ref{fig:\sceneid_ours_inf_out_uncorrupt} and \ref{fig:\sceneid_ours_inf_out_corrupt} show the results of training with artificially injected outliers, with / without corruption of the test scene as well. Finally, the corresponding results of DPESFM~\cite{Moran_2021_ICCV} can be seen in \ref{fig:\sceneid_baseline_inf_noaug}-\ref{fig:\sceneid_baseline_inf_out_corrupt}.}
    \label{fig:\sceneid}
\end{figure*}
\begin{figure*}
    \centering
    \renewcommand{\sceneid}{PortaSanDonatoBologna}
    \renewcommand{\scenename}{Porta San Donato Bologna}
    \renewcommand{\trimleft}{8}
    \renewcommand{\trimbottom}{8}
    \renewcommand{\trimright}{8}
    \renewcommand{\trimtop}{8}
    \input{figures/viz/all_in_one.tex}
    \caption{Euclidean reconstruction of \emph{\scenename{}}. \ref{fig:\sceneid_ours_inf_noaug}-\ref{fig:\sceneid_ours_inf_aug} and \ref{fig:\sceneid_ours_inf_noaug}-\ref{fig:\sceneid_ours_inf_aug} show the results of our method with and without data augmentation and bundle adjustment, while \ref{fig:\sceneid_ours_inf_out_uncorrupt} and \ref{fig:\sceneid_ours_inf_out_corrupt} show the results of training with artificially injected outliers, with / without corruption of the test scene as well. Finally, the corresponding results of DPESFM~\cite{Moran_2021_ICCV} can be seen in \ref{fig:\sceneid_baseline_inf_noaug}-\ref{fig:\sceneid_baseline_inf_out_corrupt}.}
    \label{fig:\sceneid}
\end{figure*}
\begin{figure*}
    \centering
    \renewcommand{\sceneid}{RoundChurchCambridge}
    \renewcommand{\scenename}{Round Church Cambridge}
    \renewcommand{\trimleft}{9}
    \renewcommand{\trimbottom}{9}
    \renewcommand{\trimright}{9}
    \renewcommand{\trimtop}{9}
    \input{figures/viz/all_in_one.tex}
    \caption{Euclidean reconstruction of \emph{\scenename{}}. \ref{fig:\sceneid_ours_inf_noaug}-\ref{fig:\sceneid_ours_inf_aug} and \ref{fig:\sceneid_ours_inf_noaug}-\ref{fig:\sceneid_ours_inf_aug} show the results of our method with and without data augmentation and bundle adjustment, while \ref{fig:\sceneid_ours_inf_out_uncorrupt} and \ref{fig:\sceneid_ours_inf_out_corrupt} show the results of training with artificially injected outliers, with / without corruption of the test scene as well. Finally, the corresponding results of DPESFM~\cite{Moran_2021_ICCV} can be seen in \ref{fig:\sceneid_baseline_inf_noaug}-\ref{fig:\sceneid_baseline_inf_out_corrupt}.}
    \label{fig:\sceneid}
\end{figure*}
\begin{figure*}
    \centering
    \renewcommand{\sceneid}{SmolnyCathedralStPetersburg}
    \renewcommand{\scenename}{Smolny Cathedral St Petersburg}
    \renewcommand{\trimleft}{3}
    \renewcommand{\trimbottom}{4}
    \renewcommand{\trimright}{3}
    \renewcommand{\trimtop}{3}
    \input{figures/viz/all_in_one.tex}
    \caption{Euclidean reconstruction of \emph{\scenename{}}. \ref{fig:\sceneid_ours_inf_noaug}-\ref{fig:\sceneid_ours_inf_aug} and \ref{fig:\sceneid_ours_inf_noaug}-\ref{fig:\sceneid_ours_inf_aug} show the results of our method with and without data augmentation and bundle adjustment, while \ref{fig:\sceneid_ours_inf_out_uncorrupt} and \ref{fig:\sceneid_ours_inf_out_corrupt} show the results of training with artificially injected outliers, with / without corruption of the test scene as well. Finally, the corresponding results of DPESFM~\cite{Moran_2021_ICCV} can be seen in \ref{fig:\sceneid_baseline_inf_noaug}-\ref{fig:\sceneid_baseline_inf_out_corrupt}.}
    \label{fig:\sceneid}
\end{figure*}
\begin{figure*}
    \centering
    \renewcommand{\sceneid}{SomeCathedralInBarcelona}
    \renewcommand{\scenename}{Some Cathedral in Barcelona}
    \renewcommand{\trimleft}{5}
    \renewcommand{\trimbottom}{5}
    \renewcommand{\trimright}{5}
    \renewcommand{\trimtop}{5}
    \input{figures/viz/all_in_one.tex}
    \caption{Euclidean reconstruction of \emph{\scenename{}}. \ref{fig:\sceneid_ours_inf_noaug}-\ref{fig:\sceneid_ours_inf_aug} and \ref{fig:\sceneid_ours_inf_noaug}-\ref{fig:\sceneid_ours_inf_aug} show the results of our method with and without data augmentation and bundle adjustment, while \ref{fig:\sceneid_ours_inf_out_uncorrupt} and \ref{fig:\sceneid_ours_inf_out_corrupt} show the results of training with artificially injected outliers, with / without corruption of the test scene as well. Finally, the corresponding results of DPESFM~\cite{Moran_2021_ICCV} can be seen in \ref{fig:\sceneid_baseline_inf_noaug}-\ref{fig:\sceneid_baseline_inf_out_corrupt}.}
    \label{fig:\sceneid}
\end{figure*}
\begin{figure*}
    \centering
    \renewcommand{\sceneid}{SriVeeramakaliammanSingapore}
    \renewcommand{\scenename}{Sri Veeramakaliamman Singapore}
    \renewcommand{\trimleft}{8}
    \renewcommand{\trimbottom}{8}
    \renewcommand{\trimright}{8}
    \renewcommand{\trimtop}{8}
    \input{figures/viz/all_in_one.tex}
    \caption{Euclidean reconstruction of \emph{\scenename{}}. \ref{fig:\sceneid_ours_inf_noaug}-\ref{fig:\sceneid_ours_inf_aug} and \ref{fig:\sceneid_ours_inf_noaug}-\ref{fig:\sceneid_ours_inf_aug} show the results of our method with and without data augmentation and bundle adjustment, while \ref{fig:\sceneid_ours_inf_out_uncorrupt} and \ref{fig:\sceneid_ours_inf_out_corrupt} show the results of training with artificially injected outliers, with / without corruption of the test scene as well. Finally, the corresponding results of DPESFM~\cite{Moran_2021_ICCV} can be seen in \ref{fig:\sceneid_baseline_inf_noaug}-\ref{fig:\sceneid_baseline_inf_out_corrupt}.}
    \label{fig:\sceneid}
\end{figure*}
\begin{figure*}
    \centering
    \renewcommand{\sceneid}{YuehHaiChingTempleSingapore}
    \renewcommand{\scenename}{Yueh Hai Ching Temple Singapore}
    \renewcommand{\trimleft}{9}
    \renewcommand{\trimbottom}{9}
    \renewcommand{\trimright}{9}
    \renewcommand{\trimtop}{9}
    \input{figures/viz/all_in_one.tex}
    \caption{Euclidean reconstruction of \emph{\scenename{}}. \ref{fig:\sceneid_ours_inf_noaug}-\ref{fig:\sceneid_ours_inf_aug} and \ref{fig:\sceneid_ours_inf_noaug}-\ref{fig:\sceneid_ours_inf_aug} show the results of our method with and without data augmentation and bundle adjustment, while \ref{fig:\sceneid_ours_inf_out_uncorrupt} and \ref{fig:\sceneid_ours_inf_out_corrupt} show the results of training with artificially injected outliers, with / without corruption of the test scene as well. Finally, the corresponding results of DPESFM~\cite{Moran_2021_ICCV} can be seen in \ref{fig:\sceneid_baseline_inf_noaug}-\ref{fig:\sceneid_baseline_inf_out_corrupt}.}
    \label{fig:\sceneid}
\end{figure*}

\end{document}